\documentclass[11pt]{article}          % twocolumn
\usepackage{graphicx}
\usepackage{hyperref}
\usepackage{amsmath}
\usepackage{fullpage}
\usepackage{amssymb}
\usepackage{amsfonts}
\usepackage{latexsym}

\usepackage{color}

\usepackage{soul}
\usepackage{algorithm,algorithmic}
\usepackage{latexsym}
\usepackage{setspace}

\usepackage[space,noadjust,nocompress]{cite}

\usepackage{color}
\usepackage{listings}
\definecolor{mygreen}{RGB}{28,172,0} % color values Red, Green, Blue
\definecolor{mylilas}{RGB}{170,55,241}

\usepackage[percent]{overpic}
\usepackage[labelfont=bf]{caption}
\usepackage{multirow}

\newcommand{\E}{\mathbb{E}}

\newcommand{\bfv}{ {\bf{v}}}

\newcommand{\bbR}{\mathbb{R}}

\newcommand{\bfK}{{\bf K}}
\newcommand{\nc}{n_{\mathrm{c}}}
\newcommand{\nf}{n_{\mathrm{f}}}
\newcommand{\np}{n_{\mathrm{p}}}

\renewcommand{\bfK}{{\bf K}}
\newcommand{\bfA}{{\bf A}}
\newcommand{\bfD}{{\bf D}}

\newcommand{\bfL}{{\bf L}}

\newcommand{\bfU}{{\bf U}}

\newcommand{\bfW}{{\bf W}}

\newcommand{\bfX}{{\bf X}}
\newcommand{\bfZ}{{\bf Z}}

\newcommand{\bfc}{{\bf c}}
\newcommand{\bfd}{{\bf d}}
\newcommand{\bfe}{{\bf e}}

\newcommand{\bfs}{{\bf s}}

\newcommand{\bfr}{{\bf r}}

\newcommand{\bfy}{{\bf  y}}

\newcommand{\bfu}{{\bf u}}
\newcommand{\bfz}{{\bf z}}

\DeclareMathOperator*{\argmin}{arg\,min}                   % arg min

\usepackage{mathtools}
\usepackage{url}

\usepackage{array}
\usepackage{placeins}

\hypersetup{%
    pdftitle={ADMM-Softmax : An ADMM Approach for Multinomial Logistic Regression},
    pdfauthor={S. Wu Fung, S. Tyrv\"ainen, L. Ruthotto, E. Haber},
    pdfkeywords={machine learning, numerical optimization, alternating direction method of multipliers, classification} 
    }

\title{{ADMM-Softmax} : An ADMM Approach for Multinomial Logistic Regression}

\author{Samy Wu Fung\footnotemark[2]
        \and Sanna Tyrv\"ainen\footnotemark[3]
        \and Lars Ruthotto\footnotemark[5]
        \and Eldad Haber\footnotemark[4]}

\begin{document}

\maketitle

\renewcommand{\thefootnote}{\fnsymbol{footnote}}
\footnotetext[2]{Department of Mathematics, University of California, Los Angeles, Los Angeles, CA, USA ({\tt swufung@gmail.com})}
\footnotetext[3]{Department of Mathematics, University of British Columbia, Vancouver, Canada ({\tt sannatyr@math.ubc.ca})}
\footnotetext[5]{Department of Mathematics, Emory University, Atlanta, GA, USA ({\tt lruthotto@emory.edu})}
\footnotetext[4]{Department of Earth, Ocean and Atmospheric Sciences, University of British Columbia, Vancouver, Canada ({\tt haber@eoas.ubc.ca })}

\begin{abstract}
    We present ADMM-Softmax, an alternating direction method of multipliers (ADMM) for solving multinomial logistic regression (MLR) problems.
    Our method is geared toward supervised classification tasks with many examples and features.
    It decouples the nonlinear optimization problem in MLR into three steps that can be solved efficiently.
    In particular, each iteration of ADMM-Softmax consists of a linear least-squares problem, a set of independent small-scale smooth, convex problems, and a trivial dual variable update.
    Solution of the least-squares problem can be be accelerated by pre-computing a factorization or preconditioner, and the separability in the smooth, convex problem can be easily parallelized across examples.
    For two image classification problems, we demonstrate that ADMM-Softmax leads to improved generalization compared to a Newton-Krylov, a quasi Newton, and a stochastic gradient descent method.

    \noindent\textbf{keywords:} machine learning, nonlinear optimization, alternating direction method of multipliers, classification, multinomial regression
\end{abstract}

\section{Introduction}
\label{sec:Intro}
Multinomial classification seeks to assign the most likely label from a pre-defined set of three or more classes to all examples in a dataset.
Classification is a key step in a wide range of applications such as data mining~\cite{le2018identication}, neural signal processing~\cite{parra2001higher},  bioinformatics~\cite{tsuruoka2007learning,liao2007logistic} and text analysis~\cite{taddy2015distributed}.
The process can be described mathematically as a classifier (or hypothesis function) that maps each \emph{input feature} to a vector in the unit simplex, whose components represent the predicted probabilities for each class.

In machine learning, the classifier is modeled as a fairly general parameterized function whose parameters (also called \emph{weights}) are trained using a number of examples.
Depending on the availability of labels for this data, one distinguishes between \emph{supervised learning}, where labels are available for all examples, \emph{unsupervised learning}, where only input features are provided, and \emph{semi-supervised learning}, where only parts of the examples are labeled.
Some well-known optimization problems arising in multinomial classification include multinomial logistic regression (MLR)~\cite{friedman2001elements} and support vector machines (SVMs)~\cite{melgami2004classification} for supervised learning, and k-nearest neighbors~\cite{hastie1998discriminant} for unsupervised learning. 
In this paper, we are primarily concerned with the efficient solution of the supervised learning problem arising in MLR.

When the dimensionality of the feature vector and the number of examples are large, a key challenge in MLR is the computational expense associated with solving a convex, smooth optimization problem.
In short, the MLR problem consists of minimizing a (regularized) cross-entropy loss function used to compare the classifier's outputs to the given class probabilities.
The classifier concatenates a softmax function, which maps a vector to the unit simplex, and an affine transformation whose weights are to be learned.
Since the problem is smooth and convex, many standard optimization algorithms can be used to train the classifier, e.g., steepest descent~\cite{malouf2002comparison}, inexact-Newton or quasi-Newton methods~\cite{wright1999numerical}, and perhaps the most commonly used in the machine learning community, stochastic gradient descent (SGD)~\cite{zhang2004solving, le2011on}.  
In practice, the performance of these methods can deteriorate close to the minimizer (manifesting, e.g., in a larger number of Conjugate Gradient (CG) steps  in Newton-CG shown in Fig.~\ref{fig:HessianIllCond}) and, particularly in the case of SGD, the limited potential for parallelization. 

It is important to note that MLR is not equivalent to solving the optimization problem since an effective classifier must \emph{generalize} well, i.e., it must assign accurate labels to instances not used during training.
A known problem is \emph{overfitting}, which occurs when the optimal weights fail to generalize. 
Hence, the optimization problem is not equivalent to the learning problem. 
To gauge the risk of overfitting, we partition the available training set into three subsets: the \emph{training set}, which is used in the optimization, the \emph{validation set}, which is used to gauge the generalization of the classifier and tune regularization and other parameters through cross-validation, and the \emph{test set}, which is not used at all during training but used for the final assessment. 
As we will also demonstrate in our numerical experiments, the effect of overfitting in MLR is similar to the well-known semi-convergence in ill-posed inverse problems.
This motivates the use of iterative regularization methods that have been very successful in inverse problems; see, e.g.,~\cite{kilmer2001choosing,nagy2004iterative,hansen2005rank,chung2008weighted,gazzola2019arnoldi}. 

Our contribution consists of a reformulation of the MLR optimization problem into a constrained problem and its approximate solution through the alternating direction method of multipliers (ADMM). 
Each iteration of the scheme requires the solution of a linear least-squares (LS) problem, a separable convex, smooth optimization problem, and a trivial dual update variable; see Sec.~\ref{sec:MathADMMSoftmax}. 
The LS problem arising from the weights can be efficiently solved using direct or iterative solvers~\cite{golub1996matrix,saad2003iterative}. 
Due to its separability, the smooth and convex problem arising from the second step can be solved in parallel for each example from the training set.
We also provide all codes and examples used in this paper at \url{https://github.com/swufung/ADMMSoftmax}.

We test our method on two popular image classification problems: MNIST~\cite{lecun1998mnist}, a collection of hand-written digits, and CIFAR-10~\cite{krizhevsky2009learning}, a collection of $32\times32$ color images divided into 10 classes. 
As is common in machine learning, we embed the original data into a higher-dimensional space before beginning the training process to improve the accuracy and generalization of the classifier. In particular, we propagate the MNIST dataset through a convolution layer with randomly chosen weights, which is also known as \emph{extreme learning machines} (ELM);  see Sec.~\ref{subsubsec:MNIST} and ~\cite{huang2006extreme}. 
For the CIFAR-10 dataset, we propagate the input data through a neural network which was previously trained on a similar dataset (also known as \textit{transfer learning}). The pre-trained network we choose to use is AlexNet~\cite{krizhevsky2009learning}, which was trained on the ImageNet dataset~\cite{deng2009imagenet}, a dataset similar to CIFAR-10. 

Our paper is organized as follows. In Sec.~\ref{sec:MathMR}, we phrase MLR as an optimization problem and briefly review existing approaches for its solution. In Sec.~\ref{sec:MathADMMSoftmax}, we present the mathematical formulation of the proposed ADMM-Softmax algorithm and discuss its computational costs and convergence properties.
In Sec.~\ref{sec:Experiments}, we compare our method to SGD and Newton-CG on the MNIST and CIFAR-10 dataset. Finally, we conclude our paper with a discussion in Sec.~\ref{sec:Conclusions}. 

\section{Multinomial logistic regression}

\label{sec:MathMR}
In this section, we review the mathematical formulation of multinomial logistic regression and discuss some related works. 
In training, we are given labeled data $(\bfd,\bfc) \in \bbR^{\nf} \times \Delta_{\nc}$ sampled from a typically unknown probability distribution.
Here, $\bfd$ is the feature vector, $\nf$ is the number of features (e.g., number of pixels in an image), $\bfc$ is the class label, and $\Delta_{\nc}$ denotes the unit simplex in $\bbR^{\nc}$ where $\nc$ is the number of classes.
Since the class vector, $\bfc$, belongs to the unit simplex, we can interpret it as a discrete probability distribution. 

The softmax classifier used to predict the class labels is given by 
\begin{equation}\label{eq:softmaxHypothesis}
     h_{\bfW}\left(\bfd\right) = \dfrac{\exp{\left(\bfW\bfd\right)}}{\bfe_{\nc}^\top \exp{\left(\bfW \bfd\right)}}.
\end{equation}  
Here, $h_{\bfW}\left(\bfd\right) \in \bbR^{\nc}$ is the predicted class label, $\bfW \in \bbR^{n_c \times n_f}$ is a weight matrix, $\bfe_{\nc} \in \bbR^{\nc}$ is a vector of all ones, and the exponential function is applied component-wise. 
To simplify our notation, we assume that~\eqref{eq:softmaxHypothesis} already contains a \emph{bias} term that applies a constant shift to all transformed features and leads to an affine model. 
A simple way to incorporate the bias term is by appending the feature vector with a one and adding a new column to the weight matrix. 

In the training, we seek to estimate weights $\bfW$ such that $h_{\bfW}\left(\bfd\right) \approx \bfc$ and that the predicted label (i.e. the index of the largest component of $h_{\bfW}\left(\bfd\right)$) matches the true label for all pairs $(\bfd,\bfc)$. 
To quantify this, we use the expected cross-entropy loss function to measure the discrepancy between the true probabilities, $\bfc$, and the predicted probabilities, $h_{\bfW}\left(\bfd\right)$.
In particular, we write the expected loss function as
\begin{align}
        \Phi(\bfW) &= \E \left[ -\bfc^\top \log{\left(h_\bfW\left(\bfd\right)\right)}\right] 
        = \E \left[  -\bfc^\top \log{\left(\dfrac{\exp{\left(\bfW\bfd\right)}}{\bfe_{\nc}^\top \exp{(\bfW \bfd)}}\right)} \right] \nonumber
        \\
        &= \E \left[-\bfc^\top\bfW\bfd + \left(\bfc^\top \bfe_{\nc}\right)\left(\log{\left(\bfe_{\nc}^\top \exp{\left(\bfW\bfd\right)}\right)} \right) \right] \nonumber
        \\
        &=\E \left[-\bfc^\top\bfW\bfd + \log{\left(\bfe_{\nc}^\top \exp{\left(\bfW\bfd\right)}\right)}\right], 
    \label{eq:CrossEntropy3}
\end{align}
where the we use the fact that $\bfc^\top \bfe_{\nc} = 1$ since $\bfc \in \Delta_{\nc}$.
The expected loss function consists of a linear term and log-sum-exp term and thus is convex and smooth~\cite{boyd2011distributed}.
In general, the loss couples all components of $\bfW$, which we will address with our proposed method.

To further aid generalization and avoid overfitting, it is common to add a Tikhonov regularization term and consider the following stochastic optimization problem of multinomial logistic regression
\begin{equation}
    \label{eq:stochMinProb}
    \argmin_{\bfW} \; 
    \left( \E \left[ -\bfc^\top\bfW\bfd + \log{\left(\bfe_{\nc}^\top \exp{\left(\bfW\bfd\right)}\right)} \right]
    +
    \frac{\alpha}{2} \left\| \bfL\left(\bfW - \bfW_{\rm{ref}}\right)^\top \right\|_F^2 \right)
\end{equation}
where $\bfL$ is a regularization operator that may enforce, e.g., smoothness, and $\alpha>0$ is a regularization parameter that balances the minimization of the loss and the regularity of the solution.
Note here that $\bfW$ has dimensions $\nc \times \nf$ so that each row of $\bfW$ corresponds to the features of each class.
Therefore, we transpose the weights in the regularization term so that we regularize the features rather than the classes of the weights.
The matrix $\bfW_{\rm{ref}}$ is a reference solution around which the regularizer is centered and needs to be chosen by the user.
During the training process, we monitor the effectiveness of the weights for instances in the validation set to calibrate the regularization parameter and other \emph{hyperparameters} using cross validation.
After the training, we apply the classifier to the test dataset to quantify its potential to generalize. 

Since we are primarily concerned with the splitting schemes, we focus the discussion to the quadratic regularizer. 
This choice also leads to a linear least-squares problem in our proposed scheme.
A common alternative to the quadratic regularizer is $\ell_1$ regularization, which is used to enforce sparsity of the weight matrix $\bfW$ and identify and extract the essential features in the data.
Efficient approaches exist to address the non-smoothness introduced by this regularizer, e.g., interior-point methods ~\cite{koh2007interiorpoint}, iterative shrinkage~\cite{hale2008fixed}, and hybrid algorithms~\cite{shi2010fast}; see also the survey by Yuan \emph{et al.} ~\cite{yuan2010comparison}. 
These schemes can be incorporated into ADMM-Softmax.

Optimization methods for soving~\eqref{eq:stochMinProb} can be broadly divided into two classes. 
Stochastic approximation (SA) methods such as stochastic gradient descent (SGD) and its variants, iteratively update $\bfW$ using gradient information computed with a single pair  (or a small batch of pairs) randomly chosen from the training set~\cite{RobbinsMonro1951}.
Upon a suitable choice of the step size (also called \emph{learning rate}) these methods can decrease the expected loss. 
In contrast, stochastic average approximation (SAA) schemes~\cite{KleywegtEtAl2006} approximate the expected loss with an empirical mean computed for a large batch containing $N$ randomly chosen examples $(\bfd_1,\bfc_1), \ldots, (\bfd_N,\bfc_N)$, which leads to the deterministic optimization problem
\begin{equation}
    \label{eq:minProb}
    \argmin_{\bfW} \; 
    \frac{1}{N} \sum_{j=1}^N \left[-\bfc_j^\top\bfW\bfd_j + \log{\left(\bfe_{\nc}^\top \exp{\left(\bfW\bfd_j\right)}\right)}\right]
    +
    \frac{\alpha}{2} \left\| \bfL\left(\bfW - \bfW_{\rm{ref}}\right)^\top \right\|_F^2.
\end{equation}
Since the objective function is convex and differentiable, many standard gradient-based iterative optimization algorithms are applicable.
The method proposed in this work is an SAA method, and our numerical experiments consider both SGD (an SA method) and several standard SAA methods for comparison. 
For an excellent review and discussion about the advantages and disadvantages of both approaches, we refer to~\cite{bottou2016optimization}. 

In general, the objective functions in~\eqref{eq:stochMinProb} and~\eqref{eq:minProb} are not separable, i.e., they couple all the components in $\bfW$. 
Due to the coupling, the Hessian cannot be partitioned and also changes in each iteration dependent due to the nonlinearity of the problem.
The above reasons can render their solution computationally challenging, especially for high-dimensional data, and in the case of SAA methods, a large number of samples, $N$.

\subsection{Related work}
\label{subsec:Literature}

The wide-spread use of MLR in classification problems has  led to the development of many numerical methods for its solution. 
A common thread is efforts to decouple the optimization problems~\eqref{eq:stochMinProb} and~\eqref{eq:minProb} into several subproblems that can be solved efficiently and in parallel. 

For example,~\cite{bouchard2007efficient, gopal2013distributed}, use an upper bound based on the first-order concavity property of the log-function to decouple~\eqref{eq:minProb} into $\nc$ subproblems associated with the different classes.
The new optimization problem, however, is no longer convex.
Other possible upper bounds that allow for a separable reformulation of MLR include quadratic upper bounds and a product of sigmoids. Detailed comparison of these and analytical solutions in a Bayesian setting can be found in~\cite{bouchard2007efficient}.
In~\cite{gopal2013distributed}, Gopal and Yang use the concavity bound to solve multinomial logistic regression in parallel and show that their iterative optimization of the bounded objective converges to the same optimal solution as the unbounded original model.
Related to the concavity bound, Fagan \& Iyengar~\cite{fagan2018unbiased} and Raman \emph{et al.} ~\cite{raman2016exploiting} use convex conjugate of the negative log to reformulate the problem as a double-sum that can be solved iteratively with SA methods like SGD.  

A method closely related to ours is~\cite{gopal2013distributed}, which reformulates the MLR problem~\eqref{eq:minProb}  as a constrained optimization problem that decouples the linear and nonlinear terms of the cross-entropy loss and approximately solves the problem using an ADMM approach.
Our method uses a similar splitting and parallelization scheme. 
As in~\cite{gopal2013distributed}, the existing optimization methods slightly outperform our scheme in terms of minimizing the expected loss over the training set; however, the obtained solutions of our scheme generalize better.  Both approaches are inspired by the work of Boyd \emph{et al.}~\cite{boyd2011distributed}, who solve sparse logistic regression problems parallel by splitting it across features with ADMM. 

Splitting techniques have also been applied to non-convex classification problems, e.g., the training of neural networks ~\cite{taylor2016training}.
 Here, the examples concentrate on binomial regression, which allows one to use a quadratic loss function and closed-form solutions for each iteration steps. Another related approach to train neural networks is the method of auxiliary coordinates (MAC)~\cite{carreiraperpinan2012distributed}.  In MAC, new variables are introduced to decouple the problem. Unlike ADMM, however, MAC breaks the deep nesting (i.e., function compositions)  in the objective function with the new parameters.

\section{ADMM-Softmax} \label{sec:MathADMMSoftmax}
In this section, we derive ADMM-Softmax, which is an SAA method for MLR. 
The main idea of our method is to introduce an auxiliary variable and associated constraint in~\eqref{eq:minProb} to obtain a separable objective function and then solve the problem approximately using an ADMM method.
As common, the iterations of the ADMM method break down into easy-to-solve subproblems;  see~\cite{boyd2011distributed} for an excellent introduction to ADMM. 
In our case, we obtain a regularized linear least-squares problem, a separable convex, smooth optimization problem, and a trivial dual variable update.
Efficient solvers exist for the first two subproblems.

By introducing global auxiliary variables $\bfz_1, \bfz_2,\ldots,\bfz_N \in \bbR^{\nc}$, we reformulate ~\eqref{eq:minProb}  as
\begin{equation}
    \label{eq:minGVCProb}
    \begin{split}
        \argmin_{\bfW, \bfz_1, \ldots,\bfz_N} \;\; & \frac{1}{N}\sum_{j=1}^N  \left[ -\bfc_j^\top\bfz_j + \log{\left(\bfe_{\nc}^\top \exp{\left(\bfz_j\right)}\right)} \right]
    +
    \frac{\alpha}{2} \left\| \bfL\left(\bfW - \bfW_{\rm{ref}}\right)^\top \right\|_F^2
    \\
    \text{ s.t. } \; &\bfz_j = \bfW\bfd_j, \quad j=1,\ldots,N,
    \end{split}
\end{equation}
Note that this problem is equivalent to the SAA version of MLR in~\eqref{eq:minProb}. 

To solve~\eqref{eq:minGVCProb} using ADMM-Softmax, we first consider the augmented Lagrangian
\begin{equation*}
     \begin{split}
    &\mathcal{L}_\rho\left(\bfW, \bfz_1,\ldots,\bfz_N,\bfy_1,\ldots,\bfy_N \right) =  
    \\
    &\frac{1}{N}\sum_{j=1}^N \left[ -\bfc_j^\top\bfz_j + \log{\left(\bfe_{\nc}^\top \exp{\left(\bfz_j\right)}\right)}
    + \frac{\alpha}{2} \left\Vert\bfL\left(\bfW - \bfW_{\rm ref}\right)^\top\right\Vert_F^2 
    + \bfy_j^\top \left(\bfz_j - \bfW\bfd_j\right) + \frac{\rho}{2} \left\| \bfz_j - \bfW\bfd_j \right\|_2^2 \right]
    \end{split},
\end{equation*}
where $\bfy_j \in \bbR^{\nc}$ is the Lagrange multiplier associated with the $j$th constraint, and $\rho > 0$ is a penalty parameter. 
The ADMM algorithm aims at finding the saddle point of the $\mathcal{L}_{\rho}$ by performing alternating updates. 
Denoting the values of the primal and dual variables at the $k$th iteration by $\bfW^{(k)}, \bfz_1^{(k)},\ldots,\bfz_N^{(k)},\bfy_1^{(k)},\ldots,\bfy_N^{(k)}$, respectively, the scheme consists of the following three steps
\begin{equation*}
    \begin{split}
    \bfW^{({k+1})} &= \argmin_{\bfW} \; \mathcal{L}_\rho\left(\bfW, \bfz_1^{(k)},\ldots,\bfz_N^{(k)},\bfy_1^{(k)},\ldots,\bfy_N^{(k)} \right), \\
    \bfz_j^{(k+1)} &= \argmin_{\bfz_j} \;  \mathcal{L}_\rho\left(\bfW^{(k+1)}, \bfz_1^{(k)},\ldots,\bfz_{j-1}^{(k)}, \bfz_j, \bfz_{j+1}^{(k)}\ldots,\bfz_N^{(k)},\bfy_1^{(k)},\ldots,\bfy_N^{(k)} \right), \quad
    j=1,\ldots,N,
    \\
    \bfy_j^{(k+1)} &= \bfy_j^{(k)} + \rho  \left( \bfz_j^{(k+1)} - \bfW^{(k+1)}\bfd_j \right), \;\; j=1,\ldots,N.
    \end{split}
\end{equation*}
Note that in the second step we have used the fact that the Lagrangian does not couple the variables $\bfz_{j_1}$ and $\bfz_{j_2}$ for $j_1\neq j_2$ to obtain $N$ independent subproblems that can be solved can be solved in parallel.
By introducing the scaled Lagrange multiplier $\bfu_j = (1/\rho) \bfy_j$ and dropping constant terms in the respective optimization problems, the steps simplify to
\begin{align}
    \bfW^{(k+1)} &= \argmin_{\bfW} \; \frac{\rho}{2} \sum_{j=1}^N \left(\left \|\bfz_j^{(k)} - \bfW\bfd_j + \bfu_j^{(k)}\right\|_2^2\right) + \frac{\alpha}{2} \left\|\bfL\left(\bfW - \bfW_{\rm ref}\right)^\top\right\|_F^2
    \label{eq:LSSubproblem}
    \\
    \bfz_j^{(k+1)} &= \argmin_{\bfz_j} \; -\bfc_j^\top\bfz_j + \log{\left(\bfe_{\nc}^\top \exp{\left(\bfz_j\right)}\right)} + \frac{\rho}{2} \left\| \bfz_j - \bfW^{(k+1)}\bfd_j + \bfu_j^{(k)}\right\|_2^2,
    \label{eq:z_step}
    \\
    j=1,&\ldots,N, \nonumber
    \\
    \bfu_j^{(k+1)} &= \bfu_j^{(k)} + \left( \bfz_j^{(k+1)} - \bfW^{(k+1)}\bfd_j \right), \quad j=1,\ldots,N,
    \label{eq:DualUpdate} 
    \nonumber
\end{align}
The first subproblem~\eqref{eq:LSSubproblem} is a linear least-squares problem whose coefficient matrix is independent of $k$ (see Sec.~\ref{subsec:leastsqrstep}), and the second subproblem~\eqref{eq:z_step} decouples into $N$ independent problems, each of which is a convex, smooth optimization problem and involves $n_c$ variables; see Sec.~\ref{subsec:Zstep}.

Let us note in passing that a different regularization in the original optimization problem~\eqref{eq:minProb} would only impact the least-squares subproblem in~\eqref{eq:LSSubproblem}, and the formulation of the $\bfz$-steps in~\eqref{eq:z_step} would remain unchanged. 
Therefore, one can use any existing algorithms to efficiently solve least-square problems with different types of regularizations terms. 

 \begin{algorithm}[t] % enter the algorithm environment
\caption{ADMM-Softmax} % give the algorithm a caption
\label{alg1} % and a label for \ref{} commands later in the document
    \begin{algorithmic}[1] % enter the algorithmic environment
    \REQUIRE training data $(\bfd_j, \bfc_j), \;\; j=1,\ldots,N$
    \STATE initialize $\bfW^{(1)}, \bfz_1^{(1)},\ldots, \bfz_N^{(1)}, \bfu_1^{(1)},\ldots,\bfu_N^{(1}, \rho$, and $k=1$
  \STATE compute factorization or preconditioner for the Gram matrix~\eqref{eq:NE}
    \WHILE{ $k <$ maximum number of iterations}
        \STATE compute $\bfW^{(k+1)}$ by solving least-squares problem~\eqref{eq:LSSubproblem} (use pre-factorized coefficient matrix)
        \STATE compute $\bfz_1^{(k+1)},\ldots, \bfz_N^{(k+1)}$ by solving convex smooth problem~\eqref{eq:z_step} (potentially in parallel)
        \STATE compute $\bfu_j^{(k+1)} = \bfu_j^{(k)} + \left( \bfz_j^{(k+1)} - \bfW\bfd_j^{(k+1)} \right), \quad j=1,\ldots,N,$
            \IF {stopping criteria~\eqref{eq:admmStop} is satisfied }
                    \STATE break
            \ENDIF
            \STATE  $k \leftarrow k+1$
        \ENDWHILE 
        \end{algorithmic}
\end{algorithm}

To terminate the ADMM iteration, a common stopping criteria is described in~\cite{boyd2011distributed}, where the norms of the primal and dual residuals are defined as
\begin{equation}
    \label{eq:admmTol}
    \begin{split}
    \left\| \bfr^{(k+1)} \right\|_2 = \sum_{j=1}^N \left\| \bfr_j^{(k+1)} \right\|_2 &=  \sum_{j=1}^N \left\| \bfz_j^{(k+1)} - \bfW^{(k+1)}\bfd_j \right\|_2,\\
     \text{ and } \; \quad 
    \left\| \bfs^{(k+1)} \right\|_2 = \sum_{j=1}^N\left\| \bfs_j^{(k+1)} \right\|_2 &= \sum_{j=1}^N \left\| \rho \text{ vec}\left(\left(\bfz_j^{(k+1)} - \bfz_j^{(k)}\right)\bfd_j^\top\right) \right\|_2,
    \end{split}
\end{equation}
respectively, where for any matrix $\bfX$, the operator vec($\bfX$) returns its vectorized form. The stopping criterion is satisfied when 
\begin{equation}\label{eq:admmStop}
    \begin{split}
    \left\| \bfr^{(k+1)} \right\|_2\leq \epsilon_{\mathrm{pri}}^{(k)} &= \sqrt{N \nc} \epsilon_{\mathrm{abs}} + \epsilon_{\mathrm{rel}} \text{max} 
    \left\{\left\| \bfz_1^{(k)}\right\|_2,\ldots,\left\|\bfz_N^{(k)}\right\|_2, \left\|\bfW^{(k)}\bfd_1\right\|_2,\ldots, \left\|\bfW^{(k)}\bfd_N\right\|_2\right\}\\
    \\
    \text{ and }  \; 
    \left\| \bfs^{(k+1)} \right\|_2 \leq \epsilon_{\mathrm{dual}}^{(k)} &= \sqrt{N \nc} \epsilon_{\mathrm{abs}} + \epsilon_{\mathrm{rel}}\text{max}\left\{\left\| \bfu_1^{(k)}\right\|_2,\ldots,\left\|\bfu_N^{(k)}\right\|_2\right\},
    \end{split}
\end{equation}
where $\epsilon_{\mathrm{rel}}>0$ and $\epsilon_{\mathrm{abs}}>0$ are the relative and absolute tolerances chosen by the user. For a summary of the proposed scheme, we refer to Alg.~\ref{alg1}.

\subsection{$\bfW$-update}
\label{subsec:leastsqrstep}
Letting 
\begin{equation}\label{eq:matrixForms}
    \begin{split}
        \bfZ^{(k)} &= \left[\bfz_1^{(k)} \; \bfz_2^{(k)}\ldots \bfz_N^{(k)}\right] \in \bbR^{\nc \times N},
        \\
        \bfU^{(k)} &= \left[\bfu_1^{(k)} \; \bfu_2^{(k)} \ldots \bfu_N^{(k)}\right] \in \bbR^{\nc \times N}, \text{ and }
        \\
        \bfD  &= \left[\bfd_1 \; \bfd_2 \ldots\bfd_N\right] \in \bbR^{\nf \times N},
    \end{split}
\end{equation}
we can rewrite the $\bfW$-update \eqref{eq:LSSubproblem} as
\begin{equation}
    \label{eq:weightUpdate}
    \bfW^{(k+1)} = \argmin_{\bfW} \; \frac{\rho}{2} \left\|\bfZ^{(k)} - \bfW\bfD + \bfU^{(k)}\right\|_F^2 + \frac{\alpha}{2} \left\|\bfL\left(\bfW - \bfW_{\rm ref}\right)^\top\right\|_F^2,
\end{equation}
which amounts to solving $n_c$ independent linear least-squares (i.e., one for each row in $\bfW$). This is equivalent to solving normal equations
\begin{equation}
    \begin{split}
      \bfA_{\alpha, \rho} \bfW^\top &= \rho \bfD\left(\bfZ + \bfU\right)^\top + \alpha \bfL\bfL^\top \bfW_{\rm ref}^\top, 
    \end{split}
    \label{eq:NE}
 \end{equation} 
where $\bfA_{\alpha, \rho} = \left(\rho \bfD\bfD^\top + \alpha \bfL\bfL^\top\right)$. Noting that the coefficient matrix in~\eqref{eq:NE} is not iteration-dependent, depending on the number of features, the matrix can be factorized once (e.g., using Cholesky applied to the normal equations or a thin-QR applied to the original least-squares problem~\cite{golub1996matrix}) and its inverse can be quickly applied, leading to trivial solves throughout the optimization. 
We also note that this is only one possible approach for solving~\eqref{eq:NE}, and that large-scale problems can be addressed using iterative methods; see, e.g.,~\cite{golub1996matrix,saad2003iterative,benzi2003robust,gazzola2015krylov,gazzola2016lanczos,benzi2002preconditioning,gazzola2018ir}.
As the performance of most iterative methods can be improved using pre-conditioning, we can compute a preconditioner (e.g., incomplete Cholesky factorization) in an offline phase and re-use it in the ADMM iterations.

\subsection{$\bfz$-update}
\label{subsec:Zstep}
Each of the objective functions in the $\bfz_j$-updates in~\eqref{eq:z_step} can be written as
\begin{equation}
    \begin{split}
        \Psi(\bfz_j) = \; -\bfc_j^\top\bfz_j + \log{\left(\bfe_{\nc}^\top \exp{(\bfz_j)}\right)} + \frac{\rho}{2} \left\| \bfz_j - \bfz_{j, \rm{ref}}\right\|_2^2,
    \end{split}
    \label{eq:zSeparableMisfit}
\end{equation}
where $j=1,\ldots,N$, and $\bfz_{j, \rm{ref}} = \bfW^{(k+1)}\bfd_j - \bfu_j^{(k)}$. In our numerical experiments, we solve these $\nc$-dimensional convex, smooth optimization problems using a Newton method. 
% In our numerical experiments, we do not solve these in parallel, and stack the auxiliary variables as $\bfZ = [\bfz_1 \: \bfz_2 \: \ldots \: \bfz_N]$ so that the Hessian is block diagonal of size $\nc N \times \nc N$, with each block containing 
 % we solve these $\nc$-dimensional convex, smooth optimization problems using a Newton method since $\nc = 10$. 
The  gradient and Hessian of the objective are
\begin{equation*}
    \begin{split}
        \nabla_\bfz{_j} \Psi(\bfz_j) = -\bfc_j + \frac{\exp{\left(\bfz_j\right)}}{\bfe_{\nc}^\top \exp{\left(\bfz_j\right)}} + \rho\left(\bfz_j-\bfz_{j,\rm{ref}}\right) \in \bbR^{\nc},
    \end{split}
\end{equation*}
and
\begin{equation}
    \begin{split}
    \nabla_{\bfz_j}^2 \Psi(\bfz_j) = \frac{\text{diag}\left({\exp{\left(\bfz_j\right)}}\right) }{\bfe_{\nc}^\top \exp{\left(\bfz_j\right)}} 
    - \frac{ \exp{\left(\bfz_j\right)} \exp{\left(\bfz_j\right)^\top}}{\left(\bfe_{\nc}^\top \exp{\left(\bfz_j\right)}\right)^2} 
    + \rho\text{diag}\left(\bfz_j\right) \in \bbR^{\nc \times \nc}
    \end{split},
    \label{eq:HessZ}
\end{equation}
respectively.
Here, for a generic vector $\bfv$, the operator $\text{diag}(\bfv)$ returns a diagonal matrix with $\bfv$ along its diagonal.
For examples with many classes, the Newton direction can be approximated using an iterative solver, e.g., the preconditioned conjugate gradient method (PCG). 
In this case, we do not build the Hessian in~\eqref{eq:HessZ} explicitly; instead, we implement its action on a vector in order to save memory.
\subsection{Computational costs and convergence}
\label{subsec:costs}
% \LRnote{Can we break down the costs of each step for our examples?}
A computationally challenging step in ADMM-Softmax is solving the least-squares problem~\eqref{eq:LSSubproblem}, for which there is a myriad of efficient solvers; see, e.g.,~\cite{golub1996matrix} for an extensive review. 
Noting that the coefficient matrix, $\bfA_{\alpha, \rho}$ in~\eqref{eq:NE} is iteration-independent, it could be factorized in the off-line phase with, e.g., Cholesky or thin QR, and we can trivially solve least-squares throughout the optimization.

The $\bfz$-updates in~\eqref{eq:z_step} can be solved in independently and in parallel for each $j=1,\ldots,N$. In our experiments, we solve the $\bfz$-updates using Newton-CG. Assuming $\np$ workers are available, the cost for each Hessian matrix-vector product from~\eqref{eq:HessZ} is in the order of about $\mathcal{O}\left(\frac{N}{\np}\nc^3 \right)$ FLOPS per worker, leading to very fast computations of the global variable update. If $\np = N$, that is, if we have a worker for each example, the cost per worker is in the order of $\mathcal{O}(\nc^3)$ FLOPS. We note that the number of class labels is usually relatively small compared to the number of features (in our experiments, for instance, $\nc = 10$), making ~\eqref{eq:z_step} negligible when solved in parallel.

As for convergence, it has been shown that the ADMM algorithm converges linearly for convex problems with an existing solution regardless of the choice $\rho>0$~\cite{eckstein1992douglas}. If the subproblems~\eqref{eq:LSSubproblem} or ~\eqref{eq:z_step} are solved inexactly, ADMM still converges under additional conditions. In particular, the sequences 
\begin{align*}
\mu^{(k)} = \left\| \bfW^{(k+1)} - \bfW^{(k)} \right\|_F \quad \text{ and } \quad  \gamma_j^{(k)} = \left\| \bfz_j^{(k+1)} - \bfz_j^{(k)} \right\|_2, \quad j=1,\ldots,N,
\end{align*}
must be summable. This allows us to solve the subproblems iteratively, especially for large-scale problems. More details can be found in~\cite{eckstein1992douglas}. 
% For the problem at hand, the regularized cross-entropy loss function is convex, and by using a direct solver for~\eqref{eq:LSSubproblem} we are guaranteed convergence to the global minimum.

\begin{figure}[t]
    \centering
    \includegraphics[width=0.8\textwidth, height=1.3in]{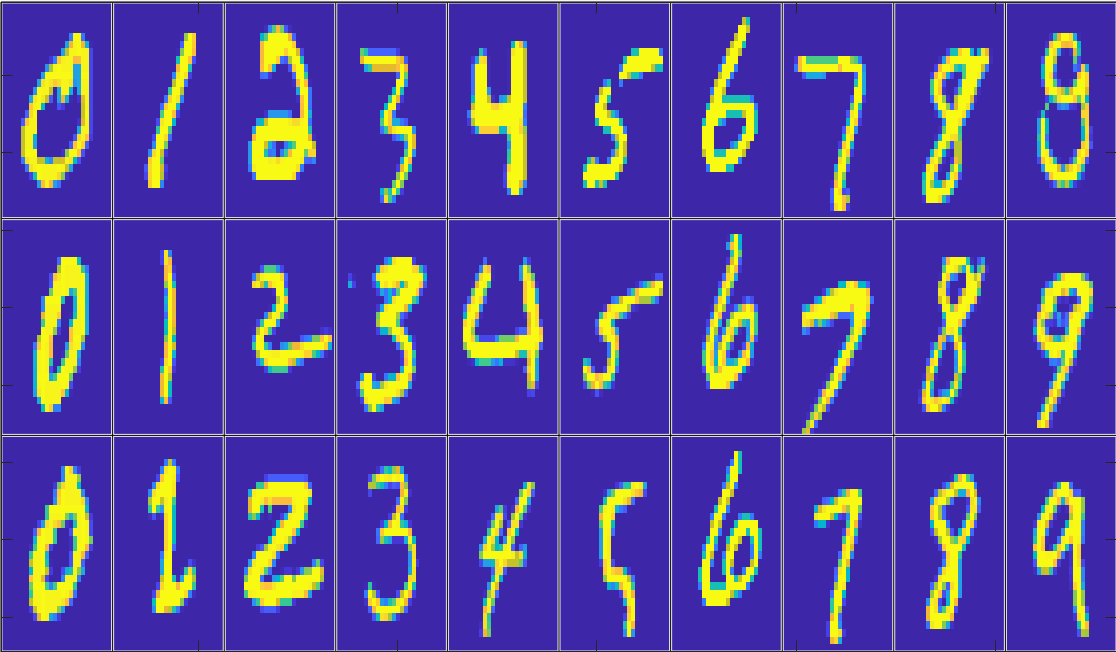}
    \caption{\small{Example of hand-written images obtained from the MNIST data set}}
    \label{fig:MNISTExample}
\end{figure}
\begin{figure}[t]
    \centering
    \begin{tabular}{cc}
    airplane
    \vspace{3mm} 
    &
    \multirow{9}{*}{\includegraphics[width=0.65\textwidth, height=2.75in]{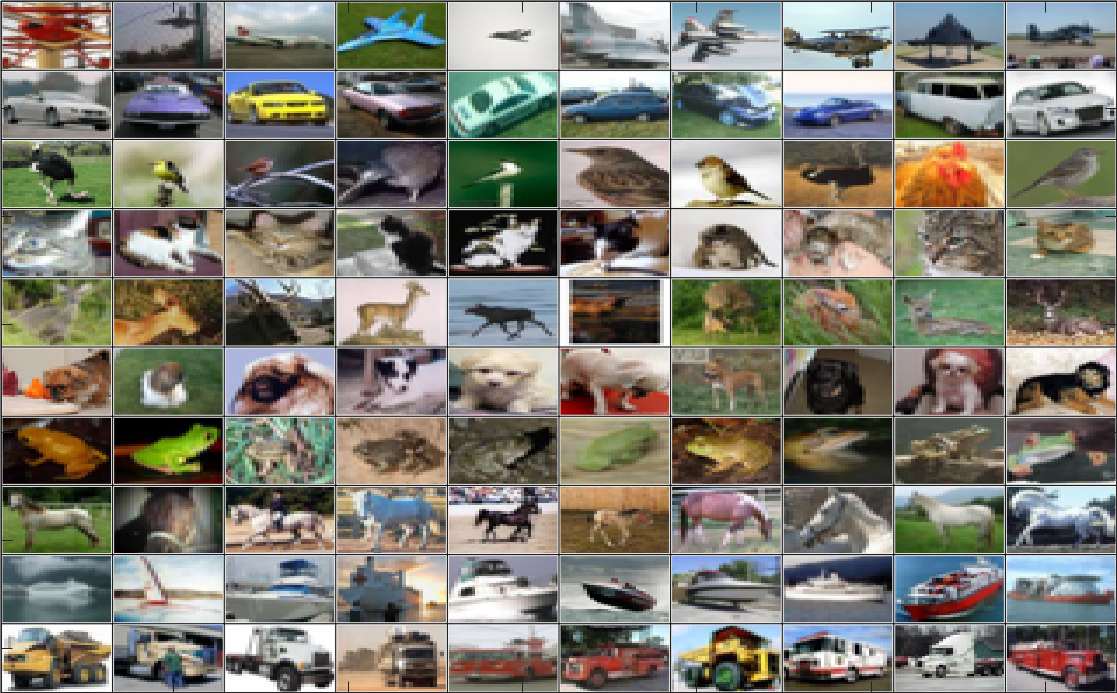}}
    \\
    automobile
    \vspace{3mm} &
    \\
    bird
    \vspace{3mm} &
    \\ 
    cat
    \vspace{3mm} & 
    \\
    deer
    \vspace{3mm} & 
    \\
    dog
    \vspace{3mm} & 
    \\
    frog
    \vspace{3mm} & 
    \\
    horse
    \vspace{3mm} &
    \\
    ship
    \vspace{3mm} & 
    \\
    truck
    \vspace{3mm} &
    \\
    \end{tabular}
    \label{CFAR10Imgs}
    \caption[Example images from the CIFAR-10 dataset]{{Example images for the CIFAR-10 dataset}}.
\end{figure}
\begin{figure}[t]
    \centering
    \begin{tabular}{ccc}
        \raisebox{0in}{
        \includegraphics[width=0.15\textwidth, height=0.8in]{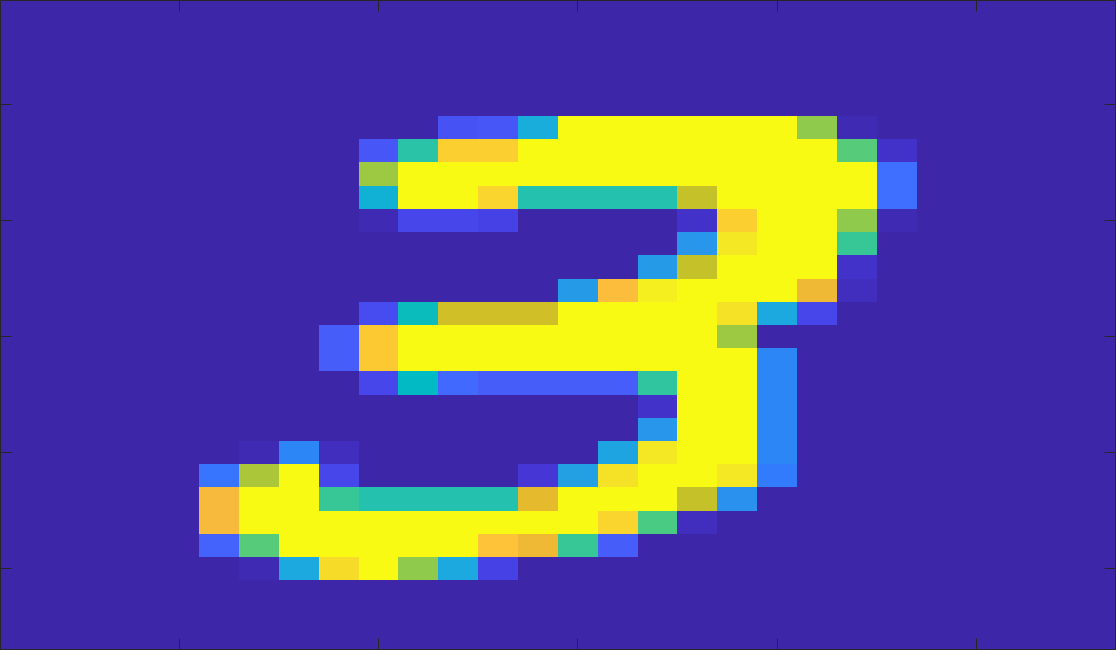}}
        &
        \raisebox{0.1in}{
        \includegraphics[width=0.25\textwidth, height=0.55in]{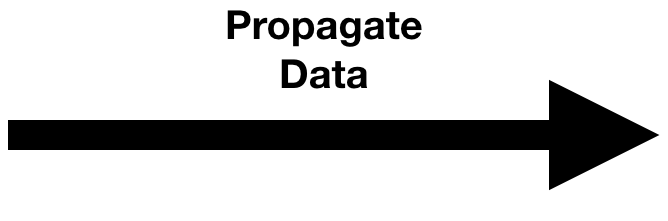}
        }
        &
        \includegraphics[width=0.2\textwidth, height=1in]{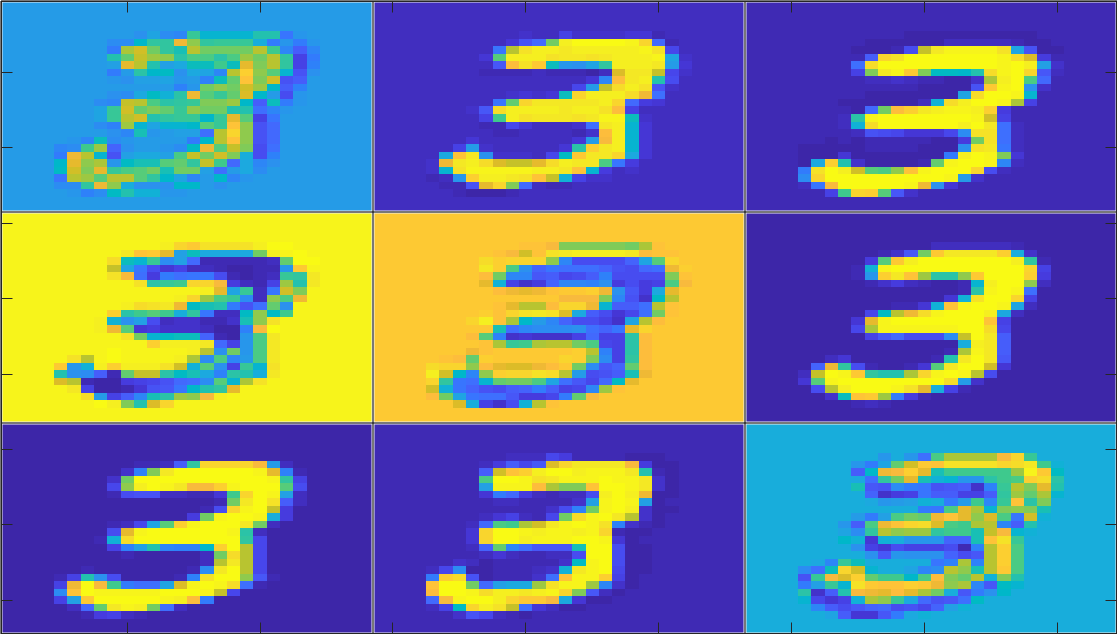}
        \\
        original image &  & propagated images
    \end{tabular}
    \caption{\small{Feature transformation of a single image from the MNIST dataset. Here, we apply a convolution layer with randomly chosen weights to increase the dimensionality of each example image increases from $ 28^2=784$ image (left) to $\nf = (3\times28)^2=7056$ (right).}}
    \label{fig:featureTransformation}
\end{figure}

\section{Numerical experiments}
\label{sec:Experiments}
In this section, we demonstrate the potential of our proposed ADMM-Softmax method using the MNIST and CIFAR-10 datasets. For both datasets, we first compare three algorithms: The SAA methods ADMM-Softmax, Newton-CG, and $\ell$-BFGS and the SA method SGD. We then study the behavior of ADMM-Softmax and its dependence on the penalty and regularization parameter.  We illustrate the challenges of Newton-CG in MLR by a comparison of the PCG performance at the first and final nonlinear iteration.  We perform our experiments in MATLAB using the Meganet deep learning package~\cite{meganet} and we provide our code at \url{https://github.com/swufung/ADMMSoftmax}.
We perform all of our experiments on a shared memory computer operating Ubuntu 14.04 with 2 Intel Xeon E5-2670 v3 2.3 GHz CPUs using 12 cores each, and a total of 128 GB of RAM. 

\subsection{Setup}
\subsubsection{MNIST} \label{subsubsec:MNIST}
The MNIST database consists of 60,000 grey-scale hand-written images of digits ranging from 0 to 9; see~\cite{lecun1998gradient, lecun1998mnist}. Here, we set 40,000 examples for training our digit-recognition system, 10,000 are used for validation, and the remaining 10,000 is used as testing data. Each image has $28 \times 28 $ pixels. A few randomly chosen examples are shown in Fig.~\ref{fig:MNISTExample}.

To improve the capacity of our multinomial regression model, we embed the original features into a higher-dimensional space obtained by a nonlinear transformation to the original variables; this procedure is also known as \textit{extreme learning machines}~\cite{huang2006extreme}. 
In particular, for each image, we obtain nine transformed images by using a convolution layer with $3\times3$ randomly chosen filters, i.e., 
\begin{equation}
    \bfd_{j, \rm{prop}} = \tanh(\bfK\bfd_j) \in \bbR^{m},
    \label{eq:fwdProp}
\end{equation}
where all 9 vertical blocks of $\bfK \in \bbR^{9 \nf \times \nf}$ are 2D convolution operators. 
Here, we assume periodic boundary conditions which render each block to be a block-circulant matrix with circulant blocks (BCCB)~\cite{hansen2006deblurring}.
The transformed features have dimensions $m = 9 \cdot \nf + 1 =  7057$. 
Expanding the features increases the effective rank of the feature matrix $\bfD$ and increase the capacity of the classifier. An illustration of the propagated features is shown in Fig.~\ref{fig:featureTransformation}.

We compare four algorithms: our proposed ADMM-Softmax, Newton-CG, $\ell$-BFGS, and SGD.
In SGD, we use Nesterov momentum with minibatch size $300$, and learning rate $l_r = 10^{-1}$. Here, we choose the learning rate and minibatch sizes by performing a grid-search on $[10^{-12}, 10^3]$ and $[1, 500]$ to maximize the performance on the validation set, respectively. The initial learning rate grid-search is done logarithmically, whereas the minibatch sizes are uniformly spaced.
In Newton-CG, we set a maximum number of $20$ inner CG iterations per Newton iteration, with CG tolerance of $10^{-2}$. In $\ell$-BFGS, we store the $10$ most recent vectors used to approximate the action of the Hessian on a vector at each iteration, and solve the inner system with the same settings as in Newton-CG.
% \LRnote{Describe the setting of lBFGS}

For the ADMM-Softmax, we perform a grid-search on $[10^{-16}, 10^3]$ and report the $\rho$ that led to the best validation accuracy - in this case, $\rho = 10^{-7}$.
To solve the LS system, we compute a QR factorization in the off-line phase, which for this experiment took about $20$ seconds. To solve~\eqref{eq:z_step}, we use the Newton-CG method from the Meganet package using a maximum of $100$ iterations with gradient norm stopping tolerance of $10^{-8}$, and initial condition $\bfz_j^{(k)}$ (i.e., \emph{warm start}). 
Since we do not parallelize step~\eqref{eq:z_step} in our implementation, our input for~\eqref{eq:z_step} is given by $\bfZ^{(k)} = [\bfz_1^{(k)},\ldots,\bfz_N^{(k)}] \in \bbR^{\nc \times N}$, and the resulting Hessian is block diagonal. As a result, we solve the inner Newton system using a maximum of $50$ inner iterations and stopping tolerance of $10^{-8}$. Note that if we solved~\eqref{eq:z_step} in parallel using $N$ workers, we would not need more than $\nc = 10$ CG iterations, as each individual Hessian would have size $\nc \times \nc$.

For all three methods, we use a discrete Laplace operator as the regularization operator $\bfL$ with $\alpha = 10^{-6}$ to enforce smoothness of the images, and set reference weights to $\bfX_{\rm ref} = \bf0$. In this case, we chose the $\alpha$ that led to the best validation accuracy for the Newton-CG method.
\begin{figure}[t]
    \centering
  \begin{tabular}{cccc}
    \includegraphics[width=0.37\textwidth]{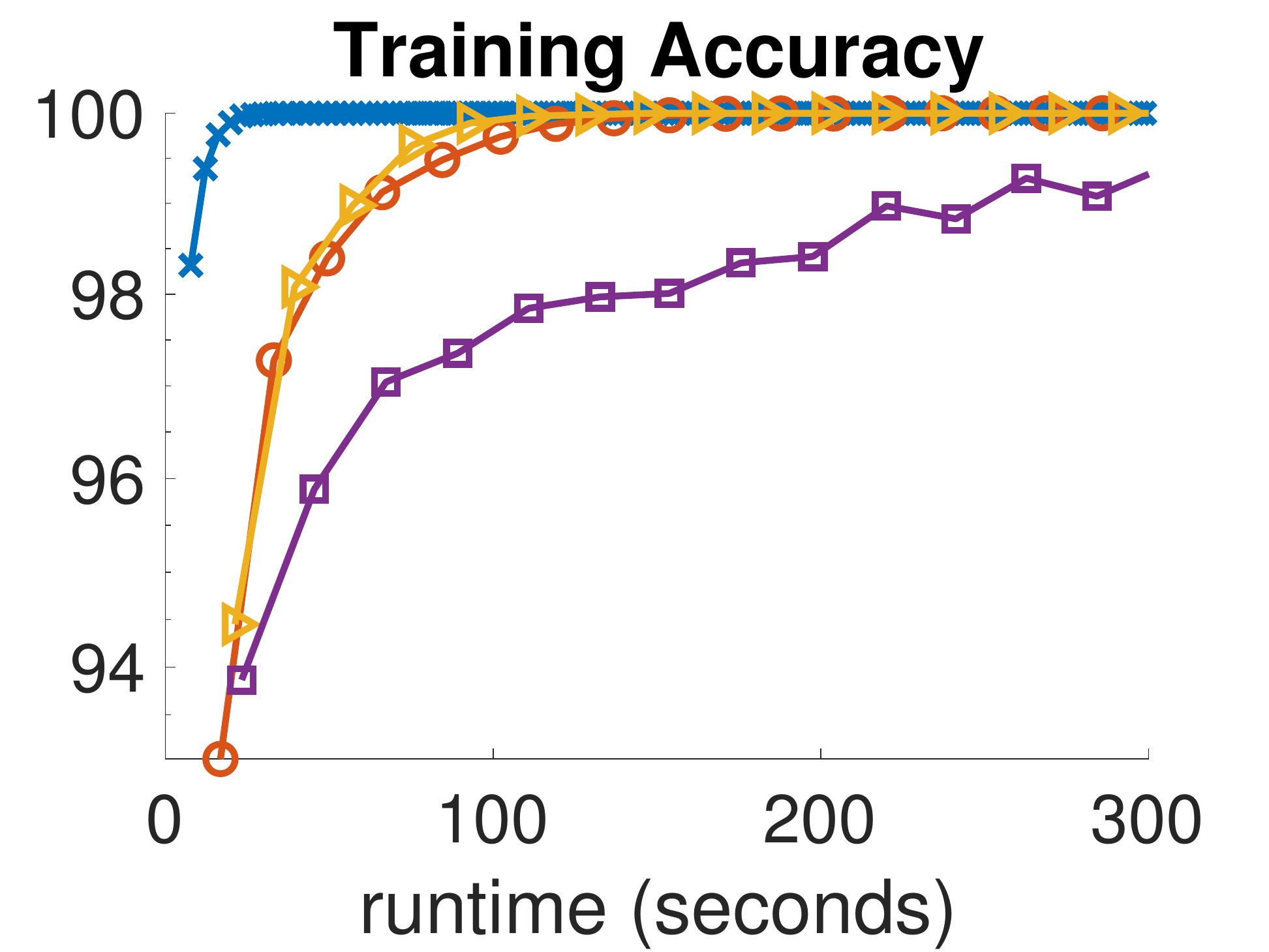}
    &
    \includegraphics[width=0.37\textwidth]{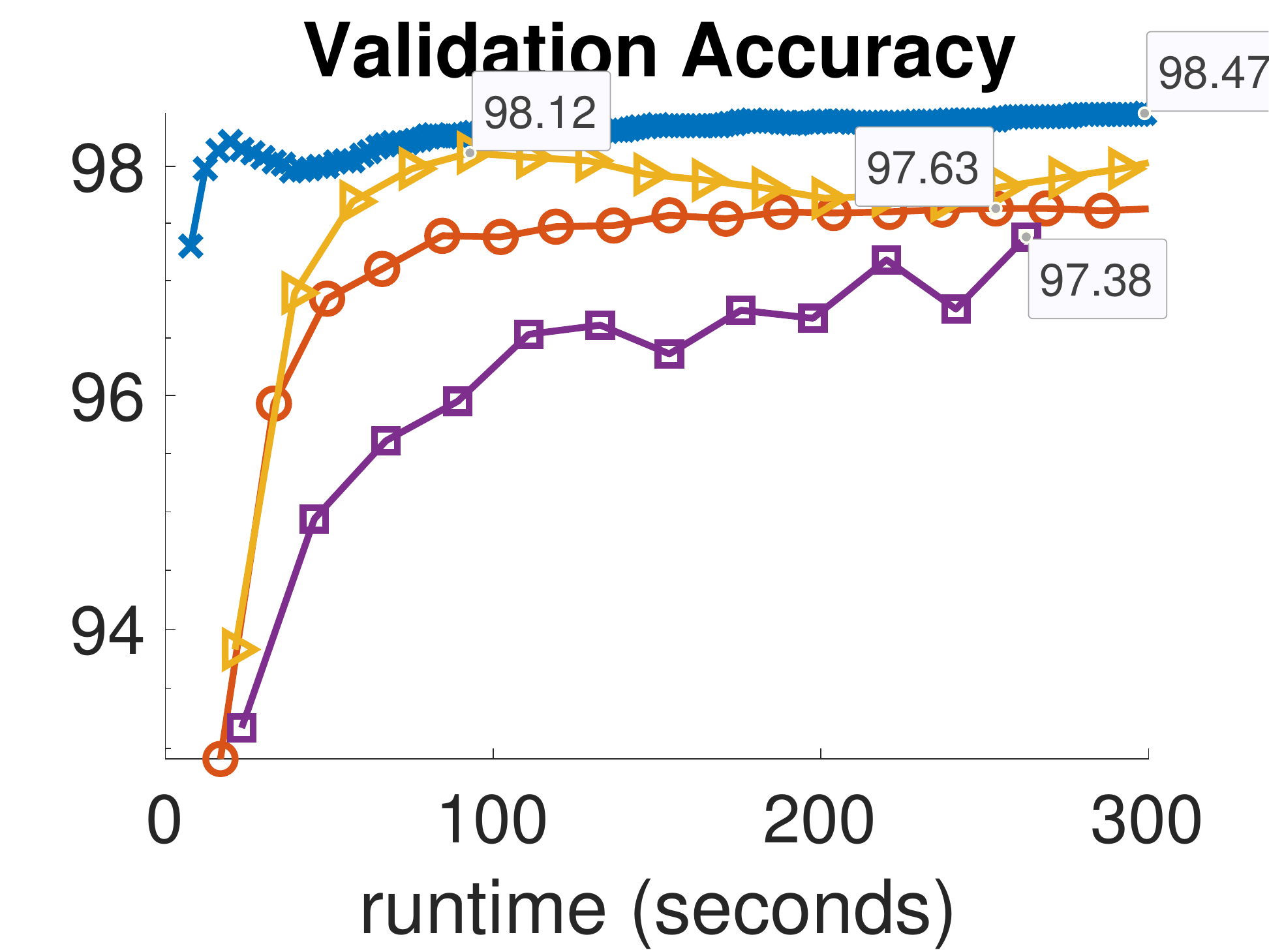}
    \\
    \includegraphics[width=0.37\textwidth]{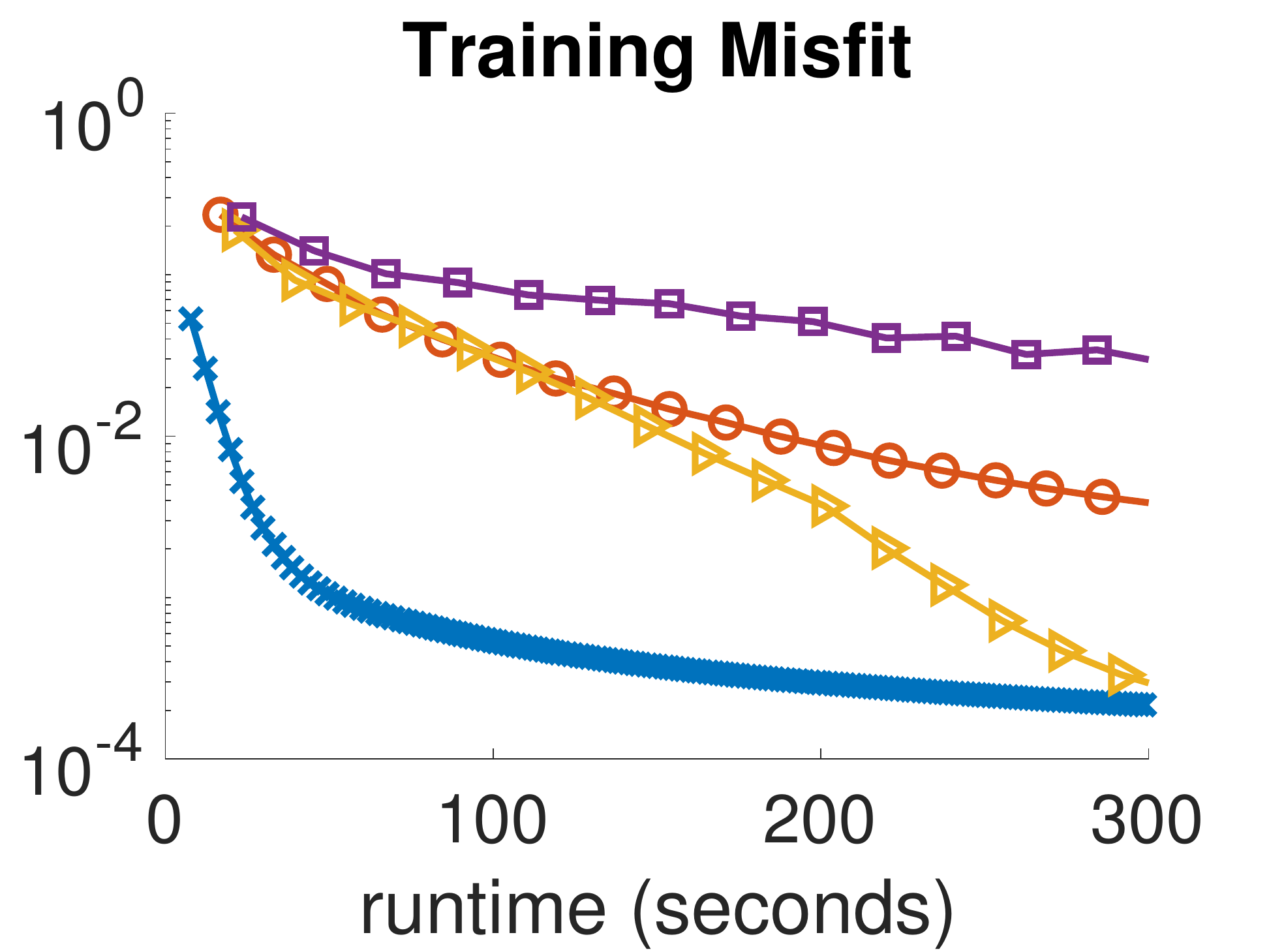}
    &
    \includegraphics[width=0.37\textwidth]{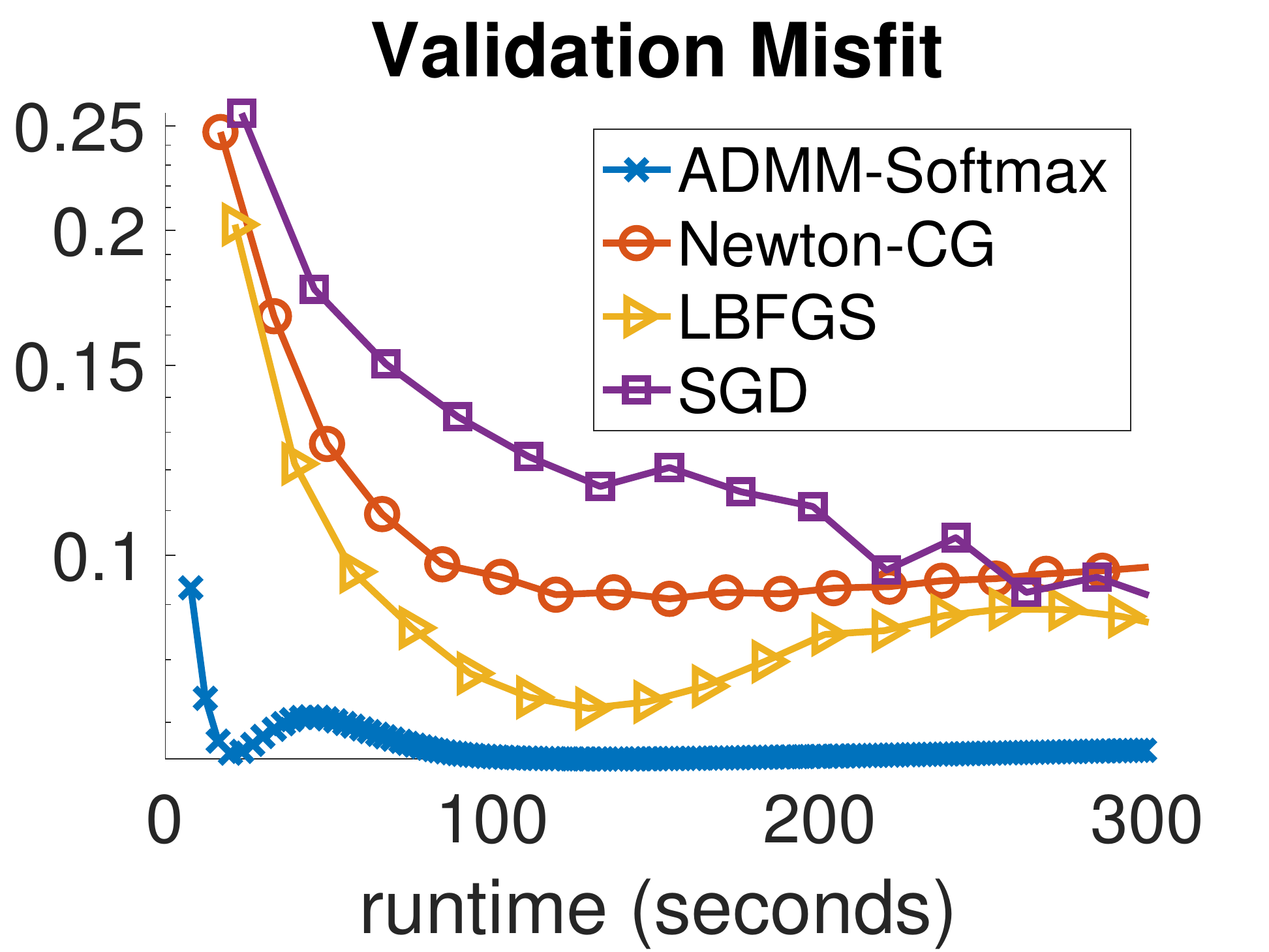}
  \end{tabular}
  \caption{For the MNIST dataset, we visualize the accuracy of the classifier (top row) and associated values of the loss functions (bottom row) computed using the training set (left column) and validation set (right column) at each iteration of the training algorithms. The x-axes show the runtime in seconds. It can be seen that the proposed ADMM-Softmax outperforms the other methods both on the training and the validation set.}
  \label{fig:MNISTConvergence}
\end{figure}
\begin{figure}[t]
    \centering
  \begin{tabular}{cccc}
    \includegraphics[width=0.37\textwidth]{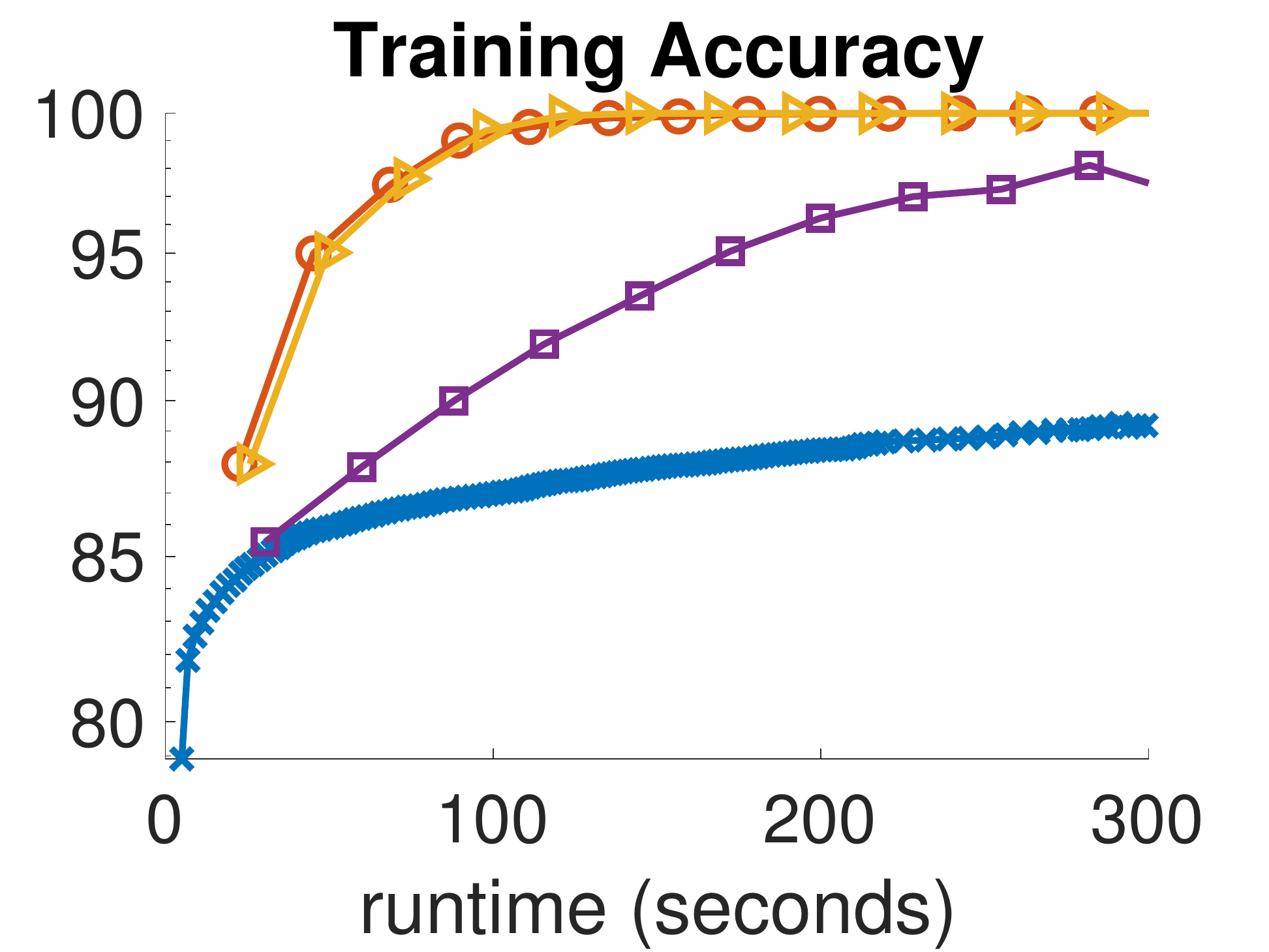}
    &
    \includegraphics[width=0.37\textwidth]{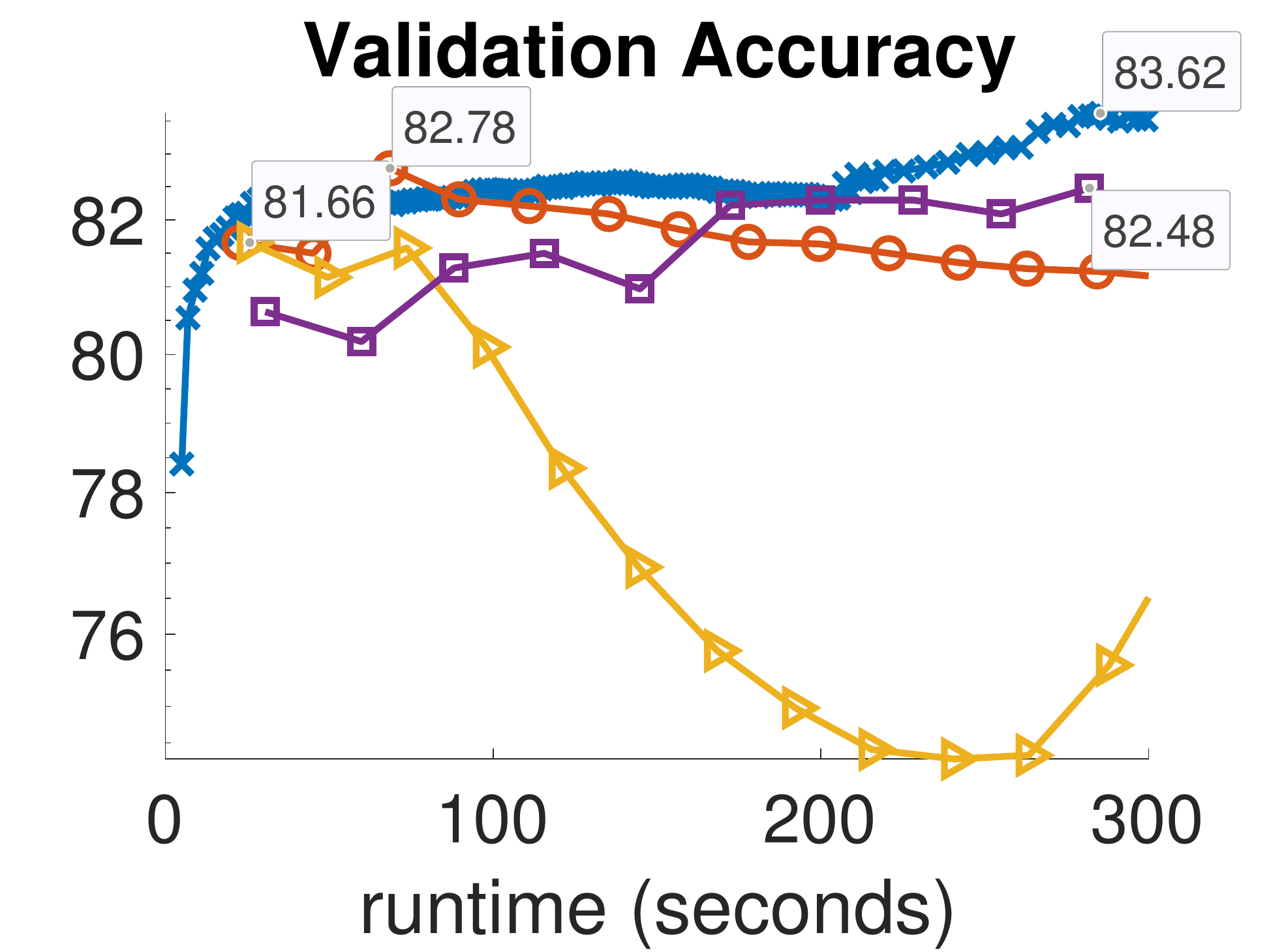}
    \\
    \includegraphics[width=0.37\textwidth]{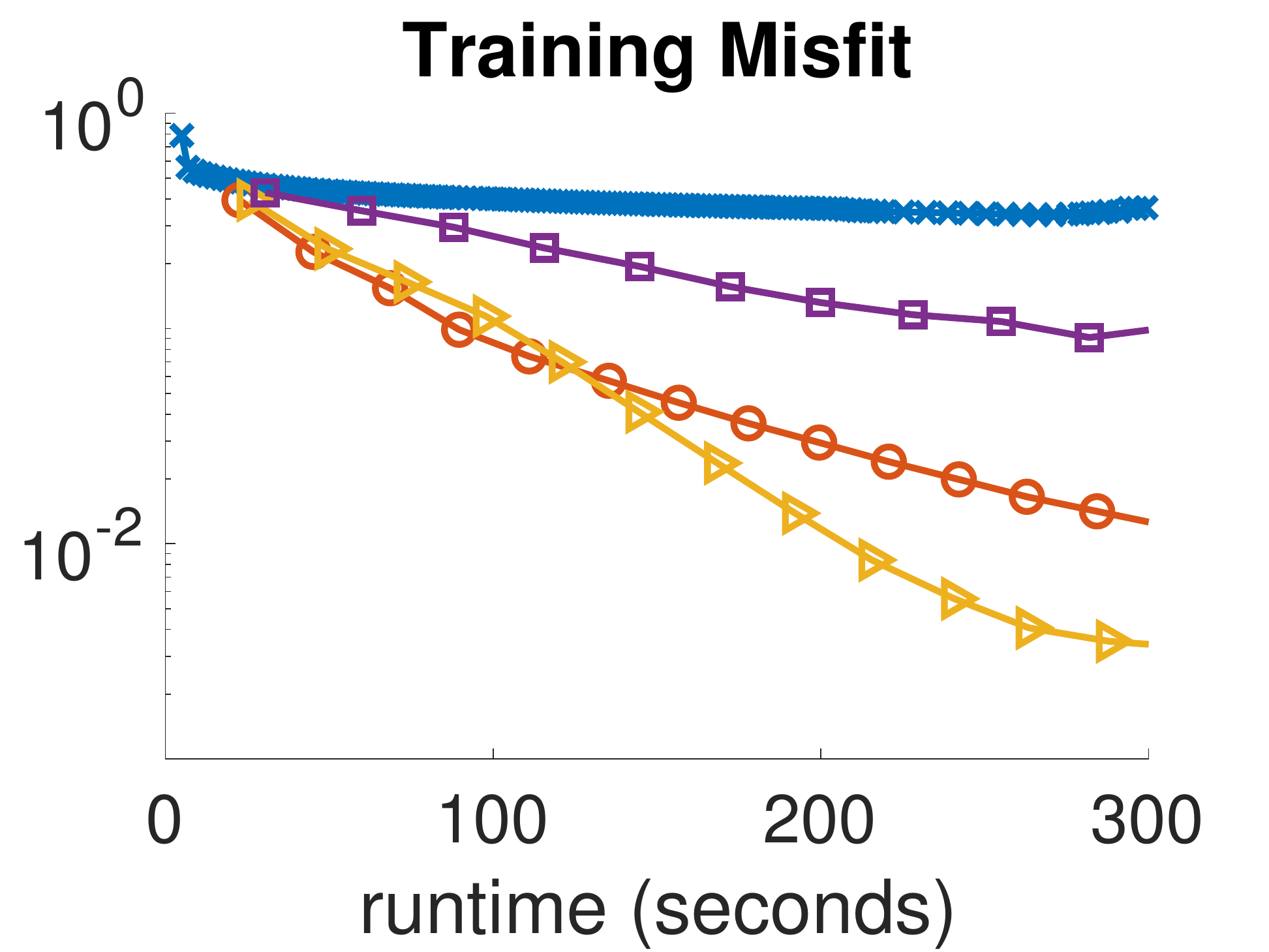}
    &
    \includegraphics[width=0.37\textwidth]{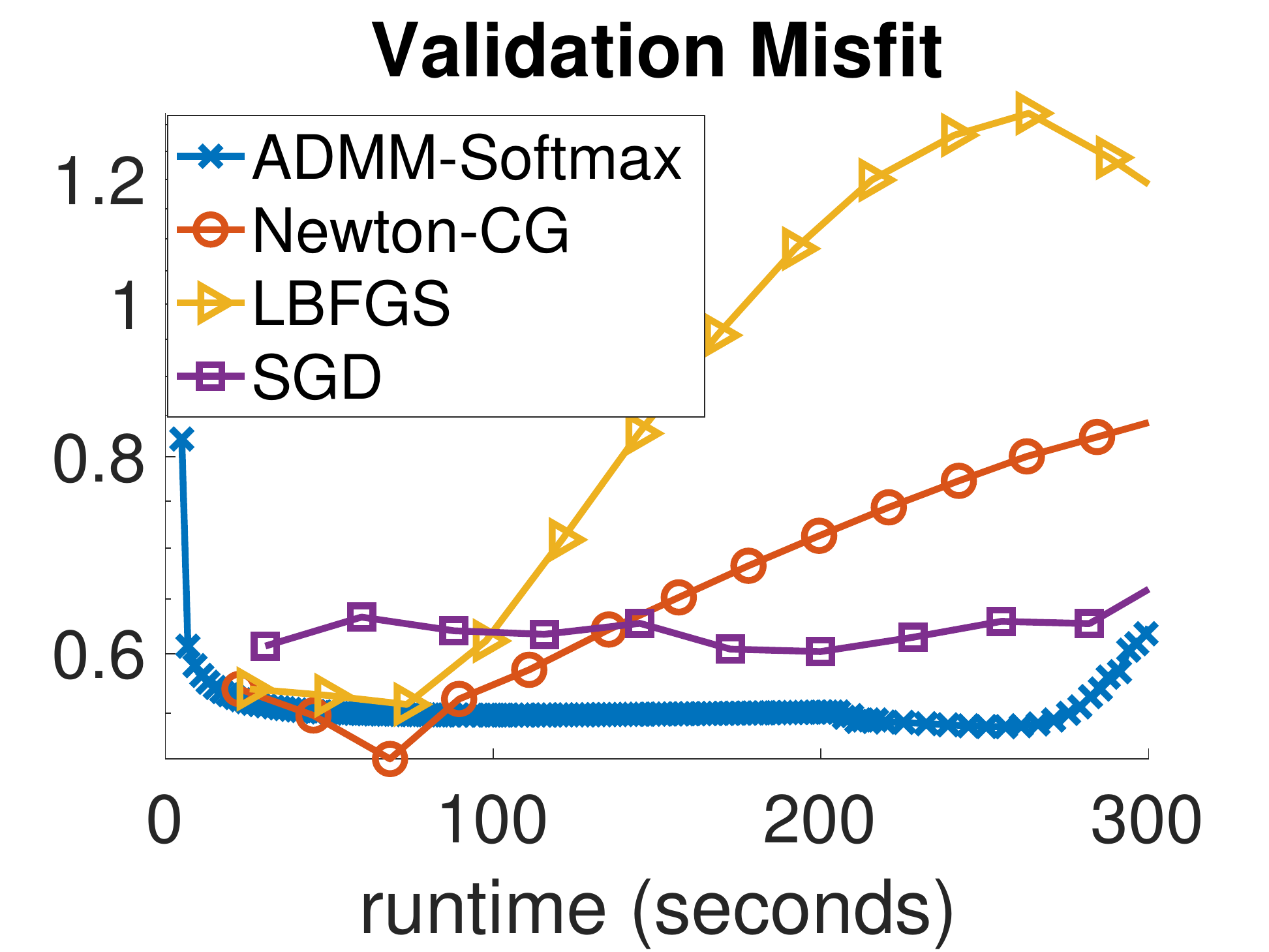}
  \end{tabular}
  \caption{For the CIFAR dataset, we visualize the accuracy of the classifier (top row) and associated values of the loss functions (bottom row) computed using the training set (left column) and validation set (right column) at each iteration of the training algorithms. The x-axes show the runtime in seconds. In this test, the ADMM-Softmax method shows inferior performance on the training dataset but leads to a slight improvement of both the loss and classification accuracy on the validation data set.}
  \label{fig:CIFAR10Convergence}
\end{figure}
\subsubsection{CIFAR-10}\label{subsubsec:CIFAR10}
The CIFAR-10 dataset~\cite{krizhevsky2009learning} consists of 60,000  RGB images of size $32 \times 32$ that are divided into 10 classes. As in MNIST, we split the data as follows: $40,000$ for training, $10,000$ as validation, and $10,000$ as testing data.
The images belong to one of the following 10 classes: airplane, automobile, bird, cat, deer, dog, frog, horse, ship, and truck.

For this dataset, we increase the feature space by propagating the input features through a neural network (AlexNet~\cite{krizhevsky2012imagenet}), which was trained on the ImageNet dataset~\cite{deng2009imagenet} using MATLAB's deep neural networks toolbox - this procedure is also known as \emph{transfer learning}. 
The main goal apart from increasing the dimensionality of the dataset is to embed the features in a way that renders them linearly separable, which is the assumption on MLR.
The main difference to the extreme learning machine that we used for MNIST is that instead of propagating through a random hidden layer as in~\eqref{eq:fwdProp}, we propagate through a network whose layers have already been trained on a similar dataset to that of CIFAR-10. 
In this case, we extract the features from the \textit{pool5} layer in AlexNet. 
For this dataset, each example $\bfd_j$ is an RGB image of dimension $32\times32\times3$ and the propagated data $\bfd_{j, \rm{prop}}$ has dimension $6\times6\times256$. 

As for the optimization, we maintain the same parameter choice as in the MNIST dataset except for the following.
For SGD, we repeat our grid-search for the learning rate and minibatch sizes as in MNIST and report the parameters that led to the best results. This leads to a smaller learning rate of $l_{r}=10^{-2}$ but the same minibatch size of $300$.
For ADMM-Softmax, we choose the penalty parameter as $\rho=8\times10^{-12}$ in the same manner as in Sec.~\ref{subsubsec:MNIST}. In this case, the propagated features no longer correspond to RGB images. 
As before, we use the Laplace operator as the regularization operator with $\bfX_{\rm{ref}} = \bf0$ and chose the $\alpha$ that led to the best results for Newton-CG, in this case, $\alpha = 10^{-1}$.

\subsection{Comparison}
\label{subsec:comparison}
In Fig.~\ref{fig:MNISTConvergence} and Fig.~\ref{fig:CIFAR10Convergence}, we show the training and validation accuracies and misfits for all algorithms on the MNIST dataset and CIFAR-10 dataset, respectively.
The former demonstrate the effectiveness of the optimization scheme, while the latter gauges the classifier's ability to generalize, which is the main concern in MLR.
Since the computational work required for each algorithm per iteration is different, we report the runtimes of each algorithm and provide an equal computational budget of 300 seconds to all methods and instances.
As can be seen, we are able to afford many more ADMM-Softmax iterations than the other algorithms within the 300-second budget (noting again that the off-line factorizations took about 20 seconds). This is because the LS-problem and $\bfz$-updates took on average $0.9$ and $0.6$ seconds, respectively, per iteration for the MNIST dataset (despite not using any parallelization). Similarly, they took $1.6$ and $0.5$ seconds on average, respectively, per iteration for the CIFAR-10 dataset. 
This is dependent on the computational platform and implementation, but we conducted a serious effort to optimize the standard methods, while not realizing the parallelization potential of ADMM-Softmax. We also provide all codes needed to replicate our experiment. 
To compare the performance of each algorithm, we pick the weights in the iteration containing the highest validation accuracy for each respective algorithm and use them to classify the testing dataset. We show these results in Tab.~\ref{tab:accuracies}. 

It is important to note that although the training process is formulated as an optimization problem, we are not necessarily solving an optimization problem, since this may not mean better generalization. Instead, we wish for an algorithm that leads to the best validation dataset.
For instance, in Fig.~\ref{fig:CIFAR10Convergence}, the Newton-CG method is the most effective algorithm at reducing the training misfit, however, this makes the algorithm overfit to the training set, leading to worse performance on the validation dataset. This can be seen in the semi-convergence behavior of the validation misfits in Fig.~\ref{fig:CIFAR10Convergence}, and is a reason for which SGD and ADMM are popular methods in the machine learning community. This situation is analogous to that of solving ill-posed inverse problems iteratively, where the objective is not necessarily choosing the parameters that best fit the (potentially noisy) data.
Finally, we note that we did not use any parallelization in any of these experiments; however, we expect that further speedups of the ADMM-Softmax method can be achieved by computing the  $\bfz$-update \eqref{eq:z_step} in parallel, particularly for larger datasets. 
\begin{table}[t]
    \centering
    \begin{tabular}{|c|c|c|c|c|c|c|c|c|}
        \hline
        & \multicolumn{4}{c|}{MNIST} & \multicolumn{4}{c|}{CIFAR-10}
        \\
        \hline
        & \multicolumn{2}{c|}{validation} & \multicolumn{2}{c|}{testing}
        & \multicolumn{2}{c|}{validation} & \multicolumn{2}{c|}{testing}
        \\
        \hline
        & misfit & accuracy & misfit & accuracy
        & misfit & accuracy & misfit & accuracy
        \\
        \hline
        ADMM-Softmax & $0.066$ & $98.47 \%$ & $0.092$ & $97.74 \%$
                     & $0.570$ & $83.62\%$ & $0.590$ & $82.90\%$
        \\
        Newton-CG    & $0.095$ & $97.63 \%$ & $0.13$ & $96.46 \%$
                     & $0.514$ & $82.78\%$  & $0.534$ & $81.58 \%$
        \\
        $\ell$-BFGS        & $0.078$ & $98.12 \%$ & $0.093$ & $97.34 \%$
                     & $0.570$ & $81.66 \%$ & $0.584$ & $81.03 \%$
        \\
        SGD          & $0.093$& $97.38 \%$ & $0.113$ & $96.56\%$
                     & $0.627$& $82.48\%$ & $0.664$ & $80.79\%$
        \\
        \hline
    \end{tabular}
    \caption{Validation and testing misfits and accuracies for ADMM-Softmax, Newton-CG, L-BFGS, and SGD (row-wise)  for the MNIST and CIFAR-10 datasets. In each case, we show the accuracy computed using the iterate with the best performance on the validation dataset. }
    \label{tab:accuracies}
\end{table}

\begin{figure}[t]
  \begin{tabular}{ccc}
        \includegraphics[width=0.31\textwidth,height=1.5in]{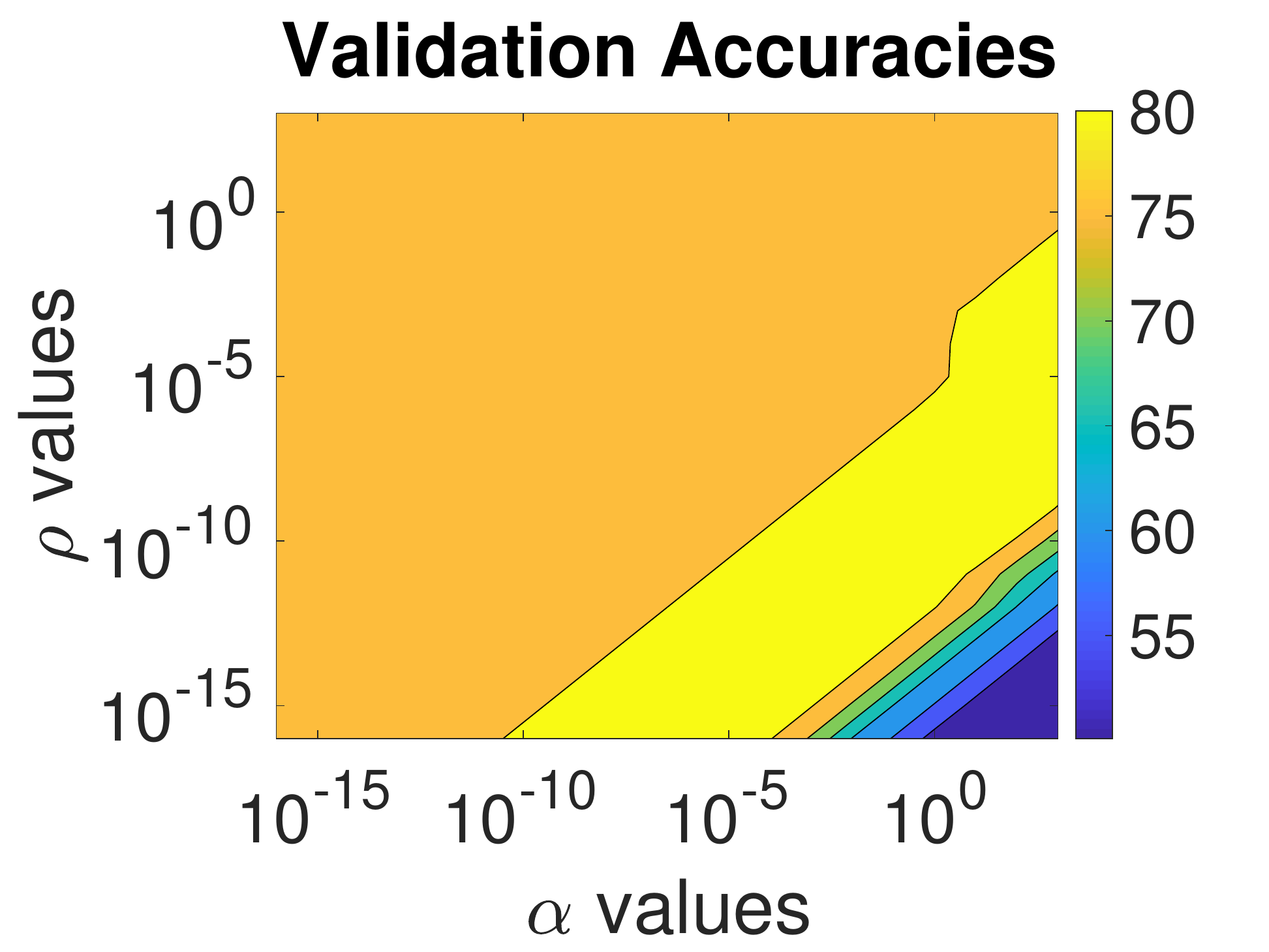}
    &
    \includegraphics[width=0.31\textwidth,height=1.5in]{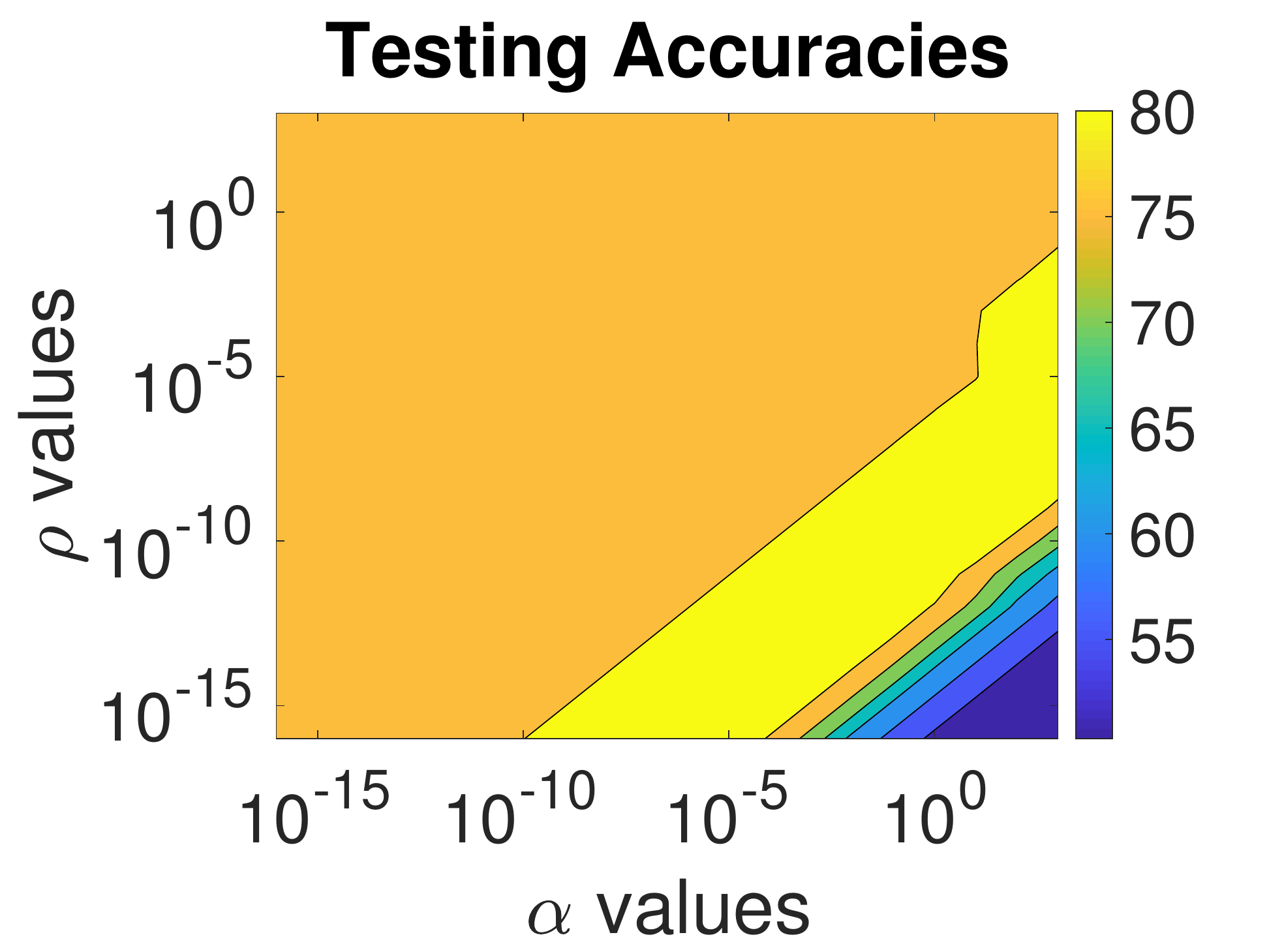}
    &
    \includegraphics[width=0.31\textwidth,height=1.5in]{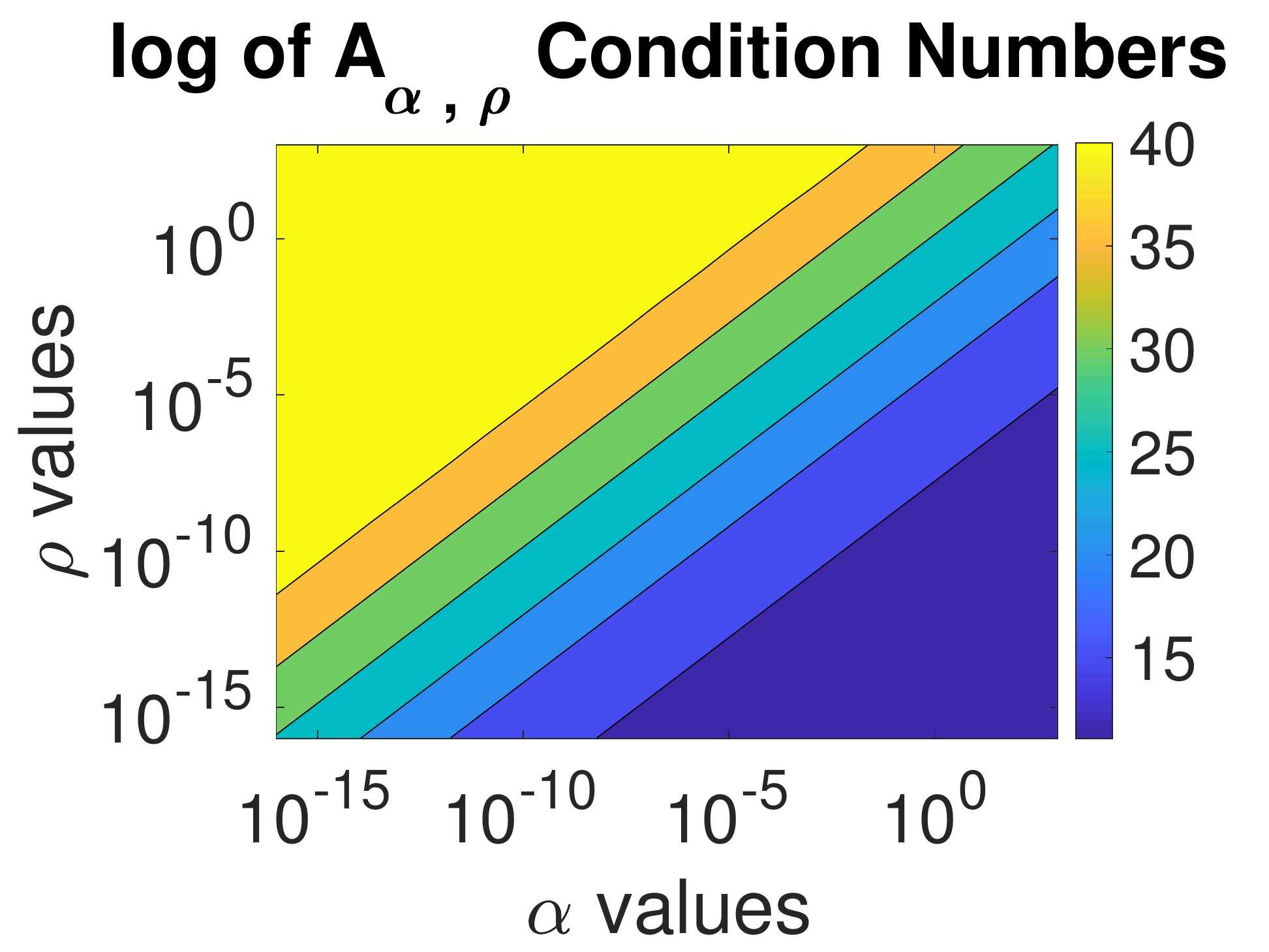}
    % \\  
    \end{tabular}
  \caption{Validation accuracies (left), and testing accuracies (right), and condition numbers of $\bfA_{\alpha, \rho}$ defined in~\eqref{eq:NE} (right) of ADMM-Softmax with respect to different $\rho$ (y-axis) and $\alpha$ (x-axis) values for the CIFAR-10 dataset. Here, $\alpha$ and $\rho$ are sampled from $[10^{-16}, 10^3]$.}
  \label{fig:paramSens}
\end{figure}

\subsection{Parameter dependence} \label{subsec:ParameterDependence}
We study the dependence of ADMM-Softmax on the penalty and regularization parameters $\rho$ and $\alpha$, where for brevity, we use only the CIFAR-10 dataset since it is the more challenging dataset; as can be seen in Sec.~\ref{subsec:comparison}.
    In Fig.~\ref{fig:paramSens}, we show the validation and testing accuracies for different values of $\rho$ and $\alpha$ sampled from $[10^{-16}, 10^{-3}]$. As in Sec.~\ref{subsec:comparison}, the weights that led to the highest validation accuracy were chosen to classify the testing dataset. The accuracy behavior is similar for the testing and validation datasets; thus, the validation dataset gives us a good indication of the generalizability of our classifier during the optimization. 
On the right, we show the condition numbers of $\bfA_{\alpha, \rho}$ (see~\eqref{eq:NE}). As expected, smaller values of $\alpha$ lead a more ill-conditioned $\bfA_{\alpha, \rho}$, however, this can be remedied with sufficiently small $\rho$.
\begin{figure}[t]
    \centering
  \begin{tabular}{cc}
        \includegraphics[width=0.37\textwidth]{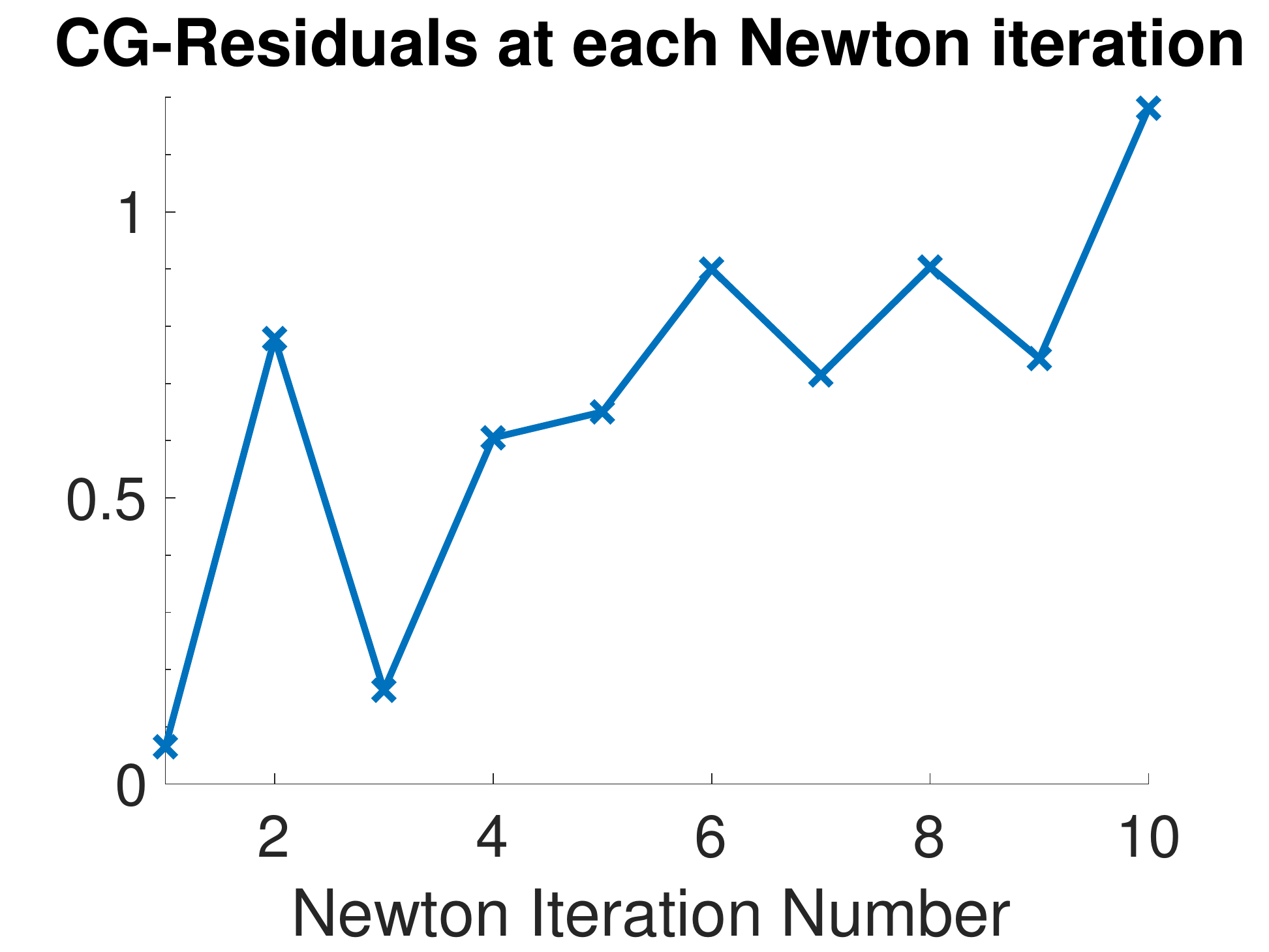}
    &
    \includegraphics[width=0.37\textwidth]{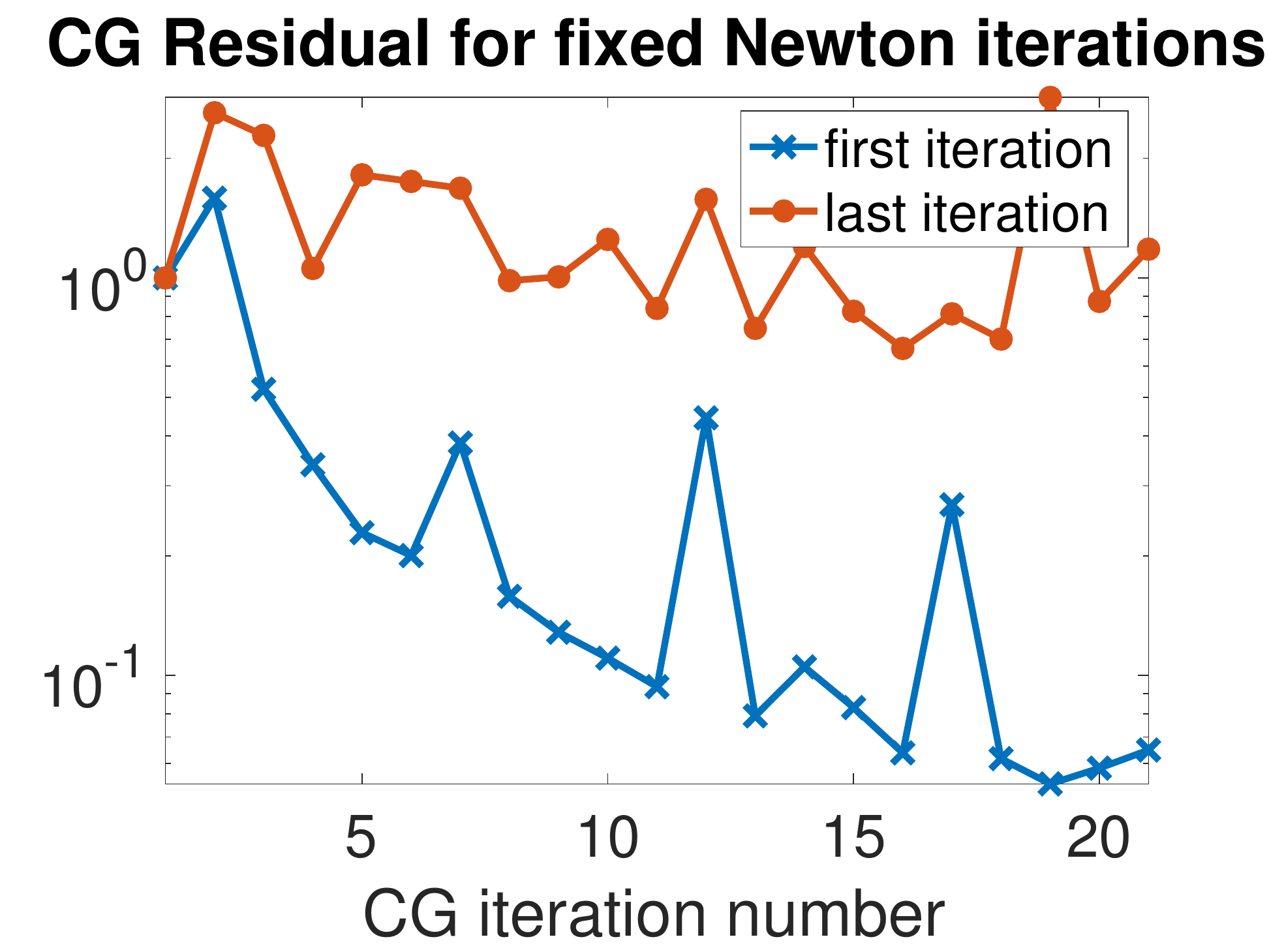}
    \end{tabular}
  \caption{We exemplarily show the deterioration of the performance of the CG iteration at the first and final Newton-CG iteration. The left subplot shows the relative residual of the solution returned by CG at each Newton iteration. The right subplot shows the relative residuals at each CG step for the first and final Newton iteration. Both plots demonstrate that the performance of CG tends to be worse at later Newton-CG iterations.}
  \label{fig:HessianIllCond}
\end{figure}
\subsection{Deterioration of Newton-CG Performance}
As a final experiment, we exemplarily show that the performance of the inner iteration of Newton-CG deteriorates as we approach the global minimum.
We have observed this phenomenon in many numerical experiments. 
Recall that the performance of CG depends mainly on the clustering of the eigenvalues of the Hessian of the objective function in~\eqref{eq:minProb}, which is iteration-dependent; for an excellent discussion of CG, see~\cite{saad2003iterative}. 
We note that the Hessians are too large for us to reliably compute a full spectral decomposition in a reasonable amount of time, so as one indicator, we plot the residuals of the CG-solver after each Newton iteration on the left of Fig.~\ref{fig:HessianIllCond}. 
% We note that the Hessians are too large for us to reliably compute a full spectral decomposition in a reasonable amount of time, so we instead report final CG-residual after each Newton iteration.
% final number of CG-residuals at each inner iteration.
The plot shows that CG is less effective in later Newton-CG iterations. 
 We also show the relative residuals for each inner CG iteration at the first and last Newton-CG iteration. Here, a decrease in performance is also evident as the last Newton iteration is quicker to stall than the first Newton iteration.
ADMM-Softmax circumvents this problem as the Hessians in the $\bfz$-update are much smaller and easier to solve.

\section{Conclusion}
\label{sec:Conclusions}
In this paper, we present ADMM-Softmax, a simple and efficient algorithm for solving multinomial logistic regression (MLR) problems arising in classification. 
To this end, we reformulate the traditional MLR problem consisting of an unconstrained optimization into a constrained optimization problem with a separable objective function. 
The new problem is solved by the alternating direction method of multipliers (ADMM), whose iteration consists of three simpler steps, i.e., a linear least-squares, a large number of independent convex, smooth optimization problems, and a trivial dual variable update. ADMM-Softmax allows the use of standard method for each of these substeps. 
In our experiments, we solve the resulting least-squares problems using a direct solver with a pre-computed factorization, and the nonlinear problems using a Newton method; see Sec.~\ref{sec:MathADMMSoftmax}.

Our method is also inspired by the successful applications of ADMM to  $\ell_1$-regularized linear inverse problems, also known as  \emph{lasso}~\cite{tibshirani1996regression,yuan2006modelselection} and \emph{basis pursuit}~\cite{chen2001atomic}. Here, ADMM  breaks the \emph{lasso} problem into two subproblems: one containing a linear least-squares problem, and the other containing a decoupled nonlinear, non-smooth term, which amends a closed-form solution given by soft thresholding~\cite{donoho1995adapting}. Our problem can be similarly divided into a linear least-squares problem and a set of decoupled smaller problems involving a nonlinear cross-entropy loss minimization. 
One distinction is that our second substep is solved using a Newton scheme.

Our numerical results show improved generalization when compared to Newton-CG and SGD for the MNIST and CIFAR-10 datasets.
Further benefits are to be expected for large datasets where parallelization is necessary. We note that better accuracies, especially for the CIFAR-10 dataset, could be achieved if we re-train the parameters of pre-trained AlexNet~\cite{yosinski2014how} (rather than keeping them fixed). To this end, our method can accelerate block-coordinate algorithms that alternate between updating the network weights and the classifier. This is a direction of our future work. Our results can be reproduced using the codes provided at \url{https://github.com/swufung/ADMMSoftmax}.

% \begin{keywords}
% machine learning, nonlinear optimization, alternating direction method of multipliers, classification, multinomial regression
% \end{keywords}
% \begin{AMS}
    % 65J22, 90C25, 49M27
% \end{AMS}

% \FloatBarrier
\section*{Acknowledgments}
This material is supported by the U.S. National Science
Foundation (NSF) through awards DMS 1522599 and DMS
1751636. We also thank the Isaac Newton Institute (INI)
for Mathematical Sciences for the support and hospitality
during the programme on generative models, parameter learning, and sparsity.

\bibliographystyle{siam}
\bibliography{Notes.bib}
% \input{MSPaper.bbl}

% % Non-BibTeX users please use
% \begin{thebibliography}{}
% %
% % and use \bibitem to create references. Consult the Instructions
% % for authors for reference list style.
% % Format for Journal Reference
% Author, Article title, Journal, Volume, page numbers (year)
% Author, Book title, page numbers. Publisher, place (year)
% \end{thebibliography}
\end{document}